\documentclass{article}
\usepackage{ijcai24}

\usepackage{times}
\usepackage{soul}
\usepackage{url}
\usepackage[hidelinks]{hyperref}
\usepackage[utf8]{inputenc}
\usepackage[small]{caption}
\usepackage{graphicx}
\usepackage{amsmath}
\usepackage{amsthm}
\usepackage{booktabs}
\usepackage{algorithm}
\usepackage{algorithmic}
\usepackage[switch]{lineno}
\usepackage{xcolor}
\usepackage{algorithm}
\usepackage{algorithmic}
\usepackage{amsthm}
\usepackage{newfloat}
\usepackage{listings}
\usepackage{amsmath}
\newtheorem{Definition}{Definition}
\usepackage{amssymb}
\usepackage{lipsum}
\usepackage{subcaption}
\usepackage{tabularx}
\usepackage{multirow}
\usepackage{adjustbox} 
\usepackage{booktabs} 
\usepackage{makecell}
\usepackage{enumitem}

\usepackage{verbatim}
\usepackage{appendix}

\newtheorem{theorem}{Theorem}
\newtheorem{lemma}{Lemma}
\newtheorem{definition}{Definition}

\newcommand{\sysname}{DCFDG}

\title{Towards Counterfactual Fairness-aware Domain Generalization in Changing Environments\thanks{This paper is supervised by Chen Zhao and Minglai Shao.}}

\author{
Yujie Lin$^1$
\and
Chen Zhao$^2$ \and
Minglai Shao$^{1}$\thanks{Corresponding author.}\and
Baoluo Meng$^3$\and
Xujiang Zhao$^{4}$\and
Haifeng Chen$^{4}$
\affiliations
$^1$School of New Media and Communication, Tianjin University, China\\
$^2$Department of Computer Science, Baylor University, USA\\ 
$^3$GE Aerospace Research,  USA\\
$^4$NEC Labs America, USA
\emails
\{linyujie\_22, shaoml\}@tju.edu.cn,
chen\_zhao@baylor.edu,
baoluo.meng@ge.com,\\
\{xuzhao, haifeng\}@nec-labs.com
}

\begin{document}

    \maketitle

    \begin{abstract}
    \label{abstract}
     Recognizing domain generalization as a commonplace challenge in machine learning, data distribution might progressively evolve across a continuum of sequential domains in practical scenarios. While current methodologies primarily concentrate on bolstering model effectiveness within these new domains, they tend to neglect issues of fairness throughout the learning process. In response, we propose an innovative framework known as \textbf{D}isentanglement for \textbf{C}ounterfactual \textbf{F}airness-aware \textbf{D}omain \textbf{G}eneralization (\sysname{}). This approach adeptly removes domain-specific information and sensitive information from the embedded representation of classification features. To scrutinize the intricate interplay between semantic information, domain-specific information, and sensitive attributes, we systematically partition the exogenous factors into four latent variables. By incorporating fairness regularization, we utilize semantic information exclusively for classification purposes. Empirical validation on synthetic and authentic datasets substantiates the efficacy of our approach, demonstrating elevated accuracy levels while ensuring the preservation of fairness amidst the evolving landscape of continuous domains.
\end{abstract}

\maketitle

\section{Introduction}
\label{sec:intro}
 The distribution shifts across sequential data domains drive the need for machine learning models with evolving domain generalization capabilities~\cite{wang2022evolving}. It requires the development of models in learning invariant representations across distinct temporal periods, consequently enhancing generalization to evolving data distributions. The temporal alignment between source and target domains \cite{zeng2023foresee} contributes to adaptive machine learning solutions, which prove indispensable in dynamic environments or evolving data streams.

As methodologies extend domain generalization to continuously evolving environments, there is a tendency to prioritize accuracy, neglecting equitable model treatment across novel domain sequences. Fairness, a significant concern in machine learning, cannot be disregarded. Sensitive features, containing protected information, include attributes like race, gender, religion, or socioeconomic status, safeguarded by ethical considerations, legal regulations, or societal norms. For instance, during the COVID-19 pandemic, systemic algorithms exhibited discrimination against African American individuals in bank loans~\cite{miller2020algorithm}. Causal models have been widely applied in machine learning to address issues related to model fairness. Structural Causal Models (SCMs)~\cite{2001Causality} provide a means of explaining machine learning model predictions. Analyzing causal graphs and paths helps understand how the model's predictions for different groups are formed, thereby identifying and addressing potential unfair factors. Simultaneously, to analyze fairness based on SCMs, a concept known as \textit{counterfactual fairness} \cite{kusner2017counterfactual} has been introduced. This concept seeks to minimize the impact on predicted values when counterfactual interventions are applied to sensitive attributes. In the context of dynamically evolving environments, we propose a framework, denoted as \textbf{D}isentanglement for \textbf{C}ounterfactual \textbf{F}airness-aware \textbf{D}omain \textbf{G}eneralization (\sysname{}), designed to address the issue of counterfactual fairness.

Our objective can be succinctly summarized as aiming to enhance the model's generalization capacity across unfamiliar domain sequences while concurrently ensuring counterfactual fairness in decision-making. Therefore, to model the relationships among sensitive attributes, domain-specific information, and semantic information, we partition the exogenous variables into four latent variables:  1) semantic information caused by sensitive attributes: $U_s$, 2) semantic information not caused by sensitive attributes: $U_{ns}$, 3) domain-specific information caused by sensitive attributes: $U_{v1}$, and 4) domain-specific information not caused by sensitive attributes: $U_{v2}$. Among these, we posit that the distribution of semantic information remains invariant across all domains, whereas the distribution of domain-specific information varies with changes in the environment. Here, the data feature $X$ is composed of two components, wherein sensitive attribute $A$ directly causes a subset of features ($X_s$), while another subset of features ($X_{ns}$) is not directly influenced by $A$ but may still exhibit correlations with it. They are encoded in the latent space as the first two exogenous variables (i.e., $U_s$ and $U_{ns}$). The advantages of this partitioning will be elucidated in the causal structure of \sysname{} (Section~\ref{sec:Causal Structure}). By employing such an approach, we skillfully disentangle domain-specific information (i.e., $U_{v1}$ and $U_{v2}$) from the embedded representation of classification features, ensuring a reduction in the impact of environmental changes on the model while concurrently upholding its decision fairness. In conclusion, our \textit{contributions} can be summarized as follows:

\begin{itemize}[leftmargin=*]
    \item We introduce a novel causal structure framework, \sysname{}, which adeptly addresses data distributions that evolve within dynamic environments and are influenced by sensitive information. 
    To the best of our knowledge, this is the first method of addressing counterfactual fairness issues in dynamic evolving environments. 
    \item We analyze the Evidence Lower Bound (ELBO) that should be considered within evolving environments. Besides, we theoretically demonstrate the rationality of \sysname{}. 
    \item Experimental results conducted on both synthetic and real-world datasets demonstrate that \sysname{} exhibits superior predictive capabilities compared to existing exogenous variable disentanglement methods, while concurrently ensuring fairness.
\end{itemize}
    
    

\section{Related Work}
\label{sec: Related Work}

\textbf{Domain Generalization in Changing Environments.} To address the generalization issues in continuously changing environments, Bai \textit{et al.}~\shortcite{bai2022temporal} involve passing the parameters of neural networks into a temporal encoder to train domain-specific parameters for each different domain. Another approach is to separately model environmental information in both features and labels, enabling the simultaneous handling of covariate shift and concept shift~\cite{qin2022generalizing}. Zeng \textit{et al.}~\shortcite{zeng2023foresee} explore aligning the data distribution in the training domain with that in an unseen domain as a means of addressing these challenges. Additionally, a classic work proposed a model-agnostic meta-learning (MAML) algorithm that learns to adapt quickly to new domains, demonstrating its effectiveness in few-shot domain generalization~\cite{finn2017model}. Building upon this work, Zhao \textit{et al.} \shortcite{zhao2021fairness,zhao2022adaptive,zhao2023towards} introduces a method that incorporates fairness considerations. 
\textbf{Counterfactual Fairness with Variational Autoencoder.}
Consider $X$, $A$, $Y$, and $U$ as data features, sensitive attributes, classification labels, and exogenous variables, respectively. Conditional Variational Autoencoder (CVAE) \cite{sohn2015learning} extends this framework by incorporating additional conditional information, such as labels $Y$, during the generation process. Louizos \textit{et al.}~\shortcite{louizos2017causal} propses a causal graph. In their CEVAE, $A$ and $X$ have an indirect connection through $U$, while $A$ has both a direct and an indirect connection with $Y$ simultaneously. However, this approach embeds $A$'s information in $U$, rendering the counterfactual generation process of $p(y|\neg a,\mathbf u)$ infeasible. To address this issue, an enhanced causal graph is proposed, assuming that $X$ and $Y$ are caused by both $A$ and $U$~\cite{pfohl2019counterfactual}. It employs Maximum Mean Discrepancy to regularize the generations, effectively removing $A$'s information from $U$. Although this approach eliminates all $A$-related components from $U$, the ideal scenario should involve the removal of only the portion in $U$ that is caused by $A$, rather than all $A$-related components. Therefore, DCEVAE~\cite{kim2021counterfactual} is proposed to define $ X_s \subset  X$ as a subset of features caused by $A$ whereas $ X_{ns} \subset  X$ is the other subset of irrelevant features to the intervention.  The intervention on $A$ should be imposed on $ X_s$, and $ X_{ns}$ should be maintained in a counterfactual generation.

\section{Background}
\label{sec:background}

\subsection{Structual Causal Model and Do-operator}
\label{sec:scm}
 Structural causal models (SCMs) are widely used in causal inference to model the causal relationships among variables. An SCM consists of a directed acyclic graph (DAG) and a set of structural equations that define the causal relationships among the variables in the graph \cite{pearl2009causality,spirtes2000causation,pearl2018book}. The structural equation for an endogenous variable $V_i$ can be expressed as follows:
 {\small
\begin{equation}
{V_i} = f_{V_i}({{Pa}_{V_i}}, {U_{V_i}})
\label{eqn:SCM}
\end{equation}
}
where ${{Pa}_{V_i}}$ denotes the parent set of ${V_i}$ in the graph, and ${U_{V_i}}$ denotes the set of exogenous variables that directly affect ${V_i}$. The function $f_i$ represents the causal relationship between the parent variables and ${V_i}$.
SCMs are used to estimate causal effects and test causal hypotheses. By including sensitive variables in the graph and modeling their causal relationships with other variables, SCMs can adjust for sensitive and produce unbiased estimates of causal effects \cite{hernan2018causal}.

\par
\textbf{Interventions on SCMs}
 involve changing the value of a variable to a specified value. This can be represented mathematically using the do-operator, denoted by ${do}({V_i}=v)$. The do-operator separates the effect of an intervention from the effect of other variables in the system. For example, if we want to investigate the effect of drug treatment on a disease outcome, we might use the do-operator to set the value of the treatment variable to ``treated" and observe the effect on the outcome variable. In the following narrative, we will employ an alternative representation for the do-operator. For two variables: $\mathit{\hat{Y}}, A$ and given exogenous variable set $U$,
 {\small
\begin{align}
    \mathbb P\big(\hat{Y}_{A\leftarrow a}(U))=\mathbb P\big(\hat{Y}(U)|do(A=a)).
\end{align}
}
    
 \subsection{Counterfactual Fairness Problem}
 Counterfactual fairness is a concept that models fairness using causal inference tools, first introduced by \cite{kusner2017counterfactual}.
 Given a predictive problem with fairness considerations, where $A$, $X$, $Y$, and $\hat{Y}$ represent the sensitive
attributes, remaining attributes, the output of interest, and model estimation respectively. 
A SCM $\mathcal{G}:=\langle {U},{V},{F},{\mathbb P}(u) \rangle$ is given, where ${V}$ is the set of endogenous variables, $\mathbb P(v):=\mathbb P(V=v)=\sum_{\{u|f_V(V,u)=v\}} \mathbb P(u)$, and ${U}$ is the set of exogenous variables. the set of deterministic functions ${F}$ is defined in  $V_i=f_{V_i}({Pa}_{V_i}, U_{V_i})$ like Eq.\ref{eqn:SCM}. We can say predictor $\hat{Y}$ is counterfactually fair, if
 {\small
\begin{equation}
\begin{aligned}
\label{eq:counterfactual_fairness}
\mathbb P\big(\hat{Y}_{A\leftarrow a}(U)&=y|X=\mathbf{x},A=a\big)\\
&=\mathbb P\big(\hat{Y}_{A\leftarrow \neg a}(U)=y|X=\mathbf{x},A=a\big)\\
\end{aligned}
\end{equation}
}
for all $y$ and any value $\neg a$ attainable by $A$. By setting $A$ to both $a$ and $\neg a$ separately, $\hat{Y}$ evolves into two distinct variants: $\hat{Y}_{A\leftarrow a}$ and $\hat{Y}_{A\leftarrow \neg a}$. From an intuitive perspective, counterfactual fairness seeks to ensure that the values of sensitive attribute $A$ do not influence the distribution of predicted outcome $\hat{Y}$.

\subsection{Counterfactual Fairness in Evolving Environments}
We consider classification tasks where the data distribution evolves gradually with time. In training stage, we are given $T$ sequentially arriving source domains $\mathcal{S}=\{ \mathcal{D}_1,\mathcal{D}_2,...,\mathcal{D}_T \}$, where each domain $\mathcal{D}_t=\{{(\mathbf{x}_{i}^t,a_{i}^t,y_{i}^t)\}_{i=1}^{n_{t}}}$ is comprised of $n_{t}$ labeled samples for $t \in \{1,2,...,T\}$. And $\mathbf{x}$, $a$, and $y$ denote the data features, the sensitive label, and the class label respectively. The trained model will be tested on $M$ target domains $\mathcal{T}=\{ \mathcal{D}_{T+1},\mathcal{D}_{T+2},...,\mathcal{D}_{T+M}\}$, $\mathcal{D}_{t}=\{{(\mathbf x_{i}^t,a_{i}^t,y_{i}^t)\}_{i=1}^{n_{t}}}$ ($t \in \{T+1,T+2,...,T+M\}$), which are not available during training stage. For simplicity, we omit the index $i$ whenever $\mathbf{x}_{i}^t$ refers to a single data point. Our primary objective is to enhance the robustness of the model on these unseen domains to achieve higher accuracy. Meanwhile, we are also committed to ensuring classification fairness across these $M$ target domains, resulting in the following expression for Eq.\ref{eq:counterfactual_fairness}:
{\small
\begin{equation}
\begin{aligned}
\mathbb P\big(\hat{Y}_{A^t\leftarrow a^t}^t(U^t)&=y^t|X^t=\mathbf{x}^t,A^t=a^t\big)\\
&=\mathbb P\big(\hat{Y}_{A^t\leftarrow \neg a^t}^t(U^t)=y|X^t=\mathbf{x}^t,A^t=a^t\big)\nonumber\\
\end{aligned}
\end{equation}
}
for $t \in \{T+1,T+2,...,T+M\}$.

\section{Methodology}
\label{sec:method}
  In this section, we will introduce the causal structure of our model. Building upon this causal structure, we will further elaborate on the entire training process of the model, including the formulation of the loss function used.

\subsection{Causal Structure of \sysname{}}
\label{sec:Causal Structure}
\begin{figure}[!t]
    \centering
    \includegraphics[width=0.95\columnwidth]{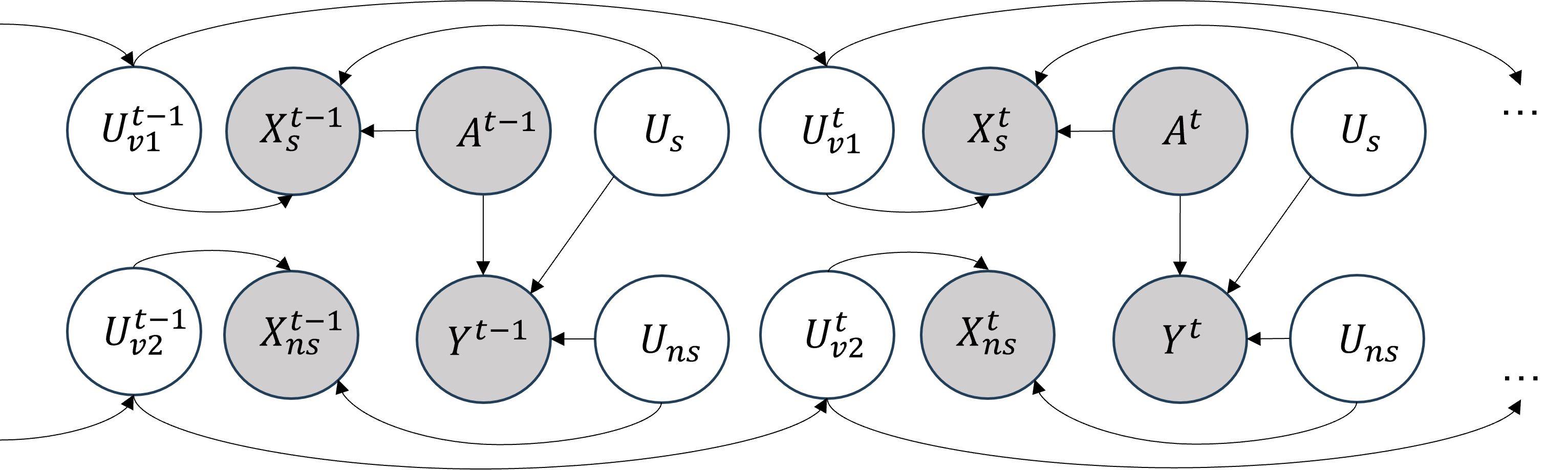}
  \caption{Causal Structure of \sysname{}. The figure depicts the causal structures across two consecutive domains, wherein, due to the gradual evolution of the environment, we posit a correlation between the environmental information of each domain and that of the preceding domain.}
  \label{fig:DGFvae}
\end{figure}

\begin{figure*}[t]
    \centering
    \includegraphics[width=0.765\linewidth]{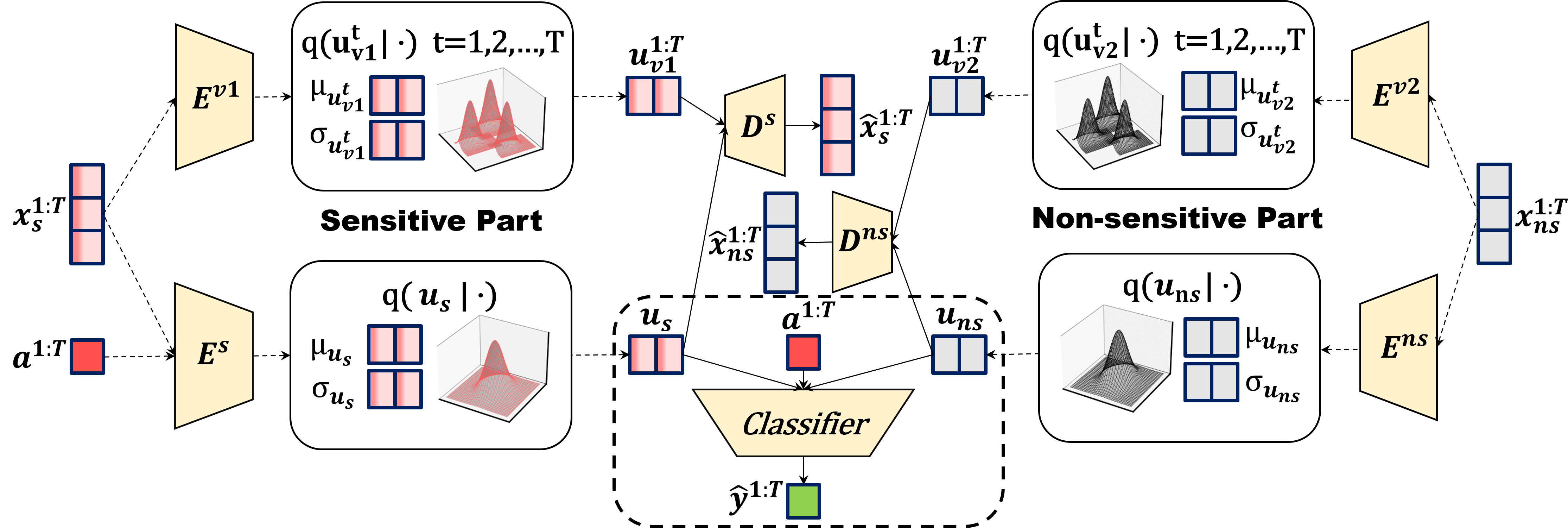}
    \caption{Network Architecture of \sysname{}. We separately decouple the environmental information $U_{v1}$ and $U_{v2}$ for $X_s$ and $X_{ns}$, and employ the adversarial loss (Section~\ref{sec:dis_loss}) to remove sensitive information from $U_s$.
    Semantic information $U_s$ and $U_{ns}$ are used for classification.
    }
    \label{fig:Network Architecture}
\end{figure*}
The causal graph depicting two consecutive domains is illustrated in Fig.~\ref{fig:DGFvae}. To achieve the counterfactual generation of $p(y|\neg a,\mathbf u)$ for intervention on $A$, it is crucial to ensure that the exogenous variable $U$ does not contain any part caused by $A$. Otherwise, there will be situations where intervention on $A$ occurs, but the information caused by $A$ in $U$ remains unchanged, leading to an erroneous generation of $y$. To address the problem, we define $X_s \subset X$ as a subset of features caused by $A$, whereas $X_{ns} \subset X$ is the other subset of irrelevant features to the intervention. This is a common method of partitioning features in the context of fairness issues~\cite{zhao2021you,grari2021fairness,kim2021counterfactual}. For instance, considering the `Sex' attribute in the Adult dataset as the sensitive attribute, we can broadly describe the characteristics of this attribute as $X_s=\{Occupation, Workclass, ...\}$, while the remaining features can be denoted as $X_{ns}$. Similarly, let's define the exogenous variables of $X_{ns}$ and $X_s$ to be $U_{ns}$ and $U_s$, respectively. We assume that $U_s$ and $U_{ns}$ are disentangled. Ideally, $U_s$ contains the portion caused by $A$, rather than the part correlated with $A$. Therefore, we need to disentangle $U_s$ from $A$. On the other hand, $U_{ns}$ contains only the part correlated with $A$ and does not require decoupling from $A$. However, in the face of a constantly changing environment, it becomes imperative to devise strategies for decoupling the domain-specific information from $X_s$ and $X_{ns}$. To simulate dynamic environments, we adopt two variables, $U_{v1}$ and $U_{v2}$, to capture the dynamic changes in the distributions of $X_s$ and $X_{ns}$ respectively, as they vary with the environments. For the domain $\mathcal{D}_t$ at timestamp t, we represent $U_{v1}$ and $U_{v2}$ as $U_{v1}^t$ and $U_{v2}^t$, respectively. 

\subsection{Network Architecture of \sysname{}}

Based on our causal graph, the corresponding neural network architecture is shown in Fig.~\ref{fig:Network Architecture}, encompassing both the inference and generation processes. During the inference stage, we employ four distinct encoders to model $q(\mathbf u_s|\mathbf x_s^t,a^t)$, $q(\mathbf u_{ns}|\mathbf x_{ns}^t)$, $q(\mathbf u_{v1}|\mathbf x_s^t)$ and $q(\mathbf u_{v2}|\mathbf x_{ns}^t)$, respectively. The prior distributions for $\mathbf u_s$ and $\mathbf u_{ns}$ follow standard normal distributions. For the environmental variable sequences $\left \{  U^t_{v1}\right \}_t^T $ and $\left \{  U^t_{v2}\right \}_t^T $, we can regard them as two temporal priors (i.e., $p(\mathbf u_{v1}^t)=p(\mathbf u_{v1}^t|\mathbf u_{v1}^{<t})$ and $p(\mathbf u_{v2}^t)=p(\mathbf u_{v2}^t|\mathbf u_{v2}^{<t})$). Hence, all the prior distributions are as follows:
{\small
\begin{align}
    &p(\mathbf u_s) = \mathcal{N}(\mathbf{0,I});\quad p(\mathbf u_{ns}) = \mathcal{N}(\mathbf{0,I});\nonumber\\
    &p(\mathbf u_{v1}^t)=p(\mathbf u_{v1}^t|\mathbf u_{v1}^{<t})=\mathcal{N}(\mu(\mathbf u_{v1}^t),\sigma^2(\mathbf u_{v1}^t));\nonumber\\
    &p(\mathbf u_{v2}^t)=p(\mathbf u_{v2}^t|\mathbf u_{v2}^{<t})=\mathcal{N}(\mu(\mathbf u_{v2}^t),\sigma^2(\mathbf u_{v2}^t)),
\end{align}}
where the distribution $p(\mathbf u_{v1}^t|\mathbf u_{v1}^{<t})$ and $p(\mathbf u_{v2}^t|\mathbf u_{v2}^{<t})$ can be encoded using recurrent neural networks such as LSTM~\cite{hochreiter1997long}. Wherein, at the initial state when $t=0$, $\mathbf u_{v1}^0$ and $\mathbf u_{v2}^0$ is initialized to $\mathbf{0}$. In the generation phase, all latent variables are fed into two distinct decoders and a classifier to reconstruct $X_s$, $X_{ns}$, and $Y$. To enhance adaptability within a dynamically changing environment, we solely utilize environment-independent semantic information to reconstruct $Y$.

\subsection{Evidence Lower Bound of \sysname{}}
For any given time point $t$ and domain $\mathcal{D}_t=\{{(\mathbf x_{i}^t,a_{i}^t,y_{i}^t)\}_{i=1}^{n_{t}}}$, we employ $U_s$ and $U_{ns}$ to capture the invariant semantic information within the distribution, while $U_{v1}^t$ and $U_{v2}^t$ are utilized to encapsulate the domain-relevant information. Analogous to the Variational Autoencoder (VAE)~\cite{kingma2013auto}, in this context, $q$ denotes the inference process, while $p$ signifies the generation process. The detailed derivation process of the ELBO for \sysname{} ~is provided in Appendix A.5. 

\textbf{Sensitive Part.} To encode representations containing sensitive information, we employ the sensitive attribute $A$ to contribute to the encoding process. Therefore, the ELBO of the sensitive part can be represented as follows:

{\small
\begin{align}
    \text{ELBO}_{s}=&\sum_{t=1}^{T}\{ \mathbb{E}_{q(\mathbf u_s|\mathbf x_s^t,a^t)q(\mathbf u_{v1}^t|\mathbf u_{v1}^{<t},\mathbf x_s^t)}\big[\log{p\left(\mathbf x_s^t|\mathbf u_s,\mathbf u_{v1}^t,a^t\right)}\big] \nonumber\\
    &-\text{KL}\big(q(\mathbf u_s|\mathbf x_s^t,a^t)||p(\mathbf u_s)\big)\nonumber\\
     &-\text{KL}\big(q(\mathbf u_{v1}^t|\mathbf u_{v1}^{<t},\mathbf x_s^t)||p(\mathbf u_{v1}^t|\mathbf u_{v1}^{<t})\big) \}.
\end{align}}
\textbf{Non-sensitive Part.} Like the sensitive part, the ELBO of the non-sensitive part can be represented as follows:
{\small
\begin{align}
    \text{ELBO}_{ns}=&\sum_{t=1}^{T}\{ \mathbb{E}_{q(\mathbf u_{ns}|\mathbf x_{ns}^t)q(\mathbf u_{v2}^t|\mathbf u_{v2}^{<t},\mathbf x_{ns}^t)}\big[\log{p\left(\mathbf x_{ns}^t|\mathbf u_{ns},\mathbf u_{v2}^t\right)}\big] \nonumber\\ 
    &-\text{KL}\big(q(\mathbf u_{ns}|\mathbf x_{ns}^t)||p(\mathbf u_{ns})\big) \nonumber \\&-
\text{KL}\big(q(\mathbf u_{v2}^t|\mathbf u_{v2}^{<t},\mathbf x_{ns}^t)||p(\mathbf u_{v2}^t|\mathbf u_{v2}^{<t})\big)\}.
\end{align}}
\textbf{Prediction Generation.} We use semantic representations and sensitive attributes for classification and the loss is:
{\small
\begin{align}
\mathcal{L}_{cla}=\sum_{t=1}^{T}\mathbb{E}_{q(\mathbf u_s|\mathbf x_s^t,a^t)q(\mathbf u_{ns}|\mathbf x_{ns}^t)}\big[\log{p\left(y^t|\mathbf u_s,\mathbf u_{ns},a^t\right)}\big].
\end{align}
}
\textbf{Final ELBO of \sysname{}.} Taking into account the three aforementioned components, we derive the final ELBO as follows:
{\small
\begin{align}
      &\quad\log p(\mathbf x_s^{1:T},\mathbf x_{ns}^{1:T},y^{1:T}|a^{1:T}) \nonumber\\
     &\geq \text{ELBO}_{s}+\text{ELBO}_{ns}+\mathcal{L}_{cla}=\text{ELBO}.
\label{eq:ELBO_jensen}
\end{align}}
During the training process, it is imperative to maximize this ELBO, consequently rendering its negative counterpart, the $-\text{ELBO }$, a constituent of the objective function.
\subsection{Counterfactual Fairness Loss of \sysname{}}The essence of counterfactual fairness lies in minimizing the impact of $A$ on the predicted value $\hat{Y}$. Therefore, for our model, if the condition:
{\small
\begin{align}
    p(\hat{y}^t|a^t,\mathbf u_s,\mathbf u_{ns})=p(\hat{y}^t|\neg a^t,\mathbf u_s,\mathbf u_{ns})
\end{align}
}
is satisfied, the model's predictions attain complete counterfactual fairness in such a case.
To earnestly achieve fairness in classification, it is imperative to augment the objective function with a fairness regularization term:
{\small
\begin{align}
\mathcal{L}_{f}=\sum_{t=1}^{T}\mathbb{E}_{q(\mathbf u_s|\mathbf x_s^t,a^t)q(y^t|x_{ns}^t)}\big[||&p(y^t|a^t, \mathbf u_s, \mathbf u_{ns})\nonumber\\-&p(y^t|\neg{a^t},\mathbf u_s, \mathbf u_{ns})||_2\big],
\end{align}}
where for the sake of simplicity, every attribute $A$ is treated as a binary variable in this paper, and $\neg a$ denotes the negation of its original value.

\subsection{Adversarial Loss of \sysname{}}
\label{sec:dis_loss}
Building upon the analysis of causal structure, $U_s$ is concurrently disentangled from both $A $ and $U_{ns}$. In other words, $U_s$ is simultaneously independent of both $A$ and $U_{ns}$ (i.e., $q(\mathbf u_s,a^t,\mathbf u_{ns})=q(\mathbf u_s)q(a^t,\mathbf u_{ns})$). Hence, the disentanglement objective is equivalent to minimizing the KL divergence between $q(\mathbf u_s,a^t,\mathbf u_{ns}) $ and $q(\mathbf u_s)q(a^t,\mathbf u_{ns})$. However, computing this KL divergence directly is infeasible, prompting us to leverage an approach akin to the one proposed in FactorVAE~\cite{kim2018disentangling}, which bears resemblance to GAN-like \cite{goodfellow2014generative} principles, to address this challenge. We begin by employing a discriminator $D$, which outputs a probability that a set of samples originates from the distribution $q(\mathbf u_s,a^t,\mathbf u_{ns}) $ rather than $q(\mathbf u_s)q(a^t,\mathbf u_{ns})$. Hence, we can approximate the KL divergence as follows using the loss function $\mathcal{L}_{TC}$ about $D$:
{\small
\begin{align}\label{eq:TC_loss}
\mathcal{L}_{TC}&=\sum_{t=1}^{T}KL\big(q(\mathbf u_s,a^t,\mathbf u_{ns})\| q(\mathbf u_s)q(a^t,\mathbf u_{ns})\big)\nonumber\\
&\approx \sum_{t=1}^{T}\mathbb{E}_{q(\mathbf u_s,a^t,\mathbf u_{ns})}\left[\log \frac{D(\mathbf u_s,a^t,\mathbf u_{ns})}{1-D(\mathbf u_s,a^t,\mathbf u_{ns})}\right].
\end{align}}
Furthermore, to train the discriminator $D$, we should maximize $\mathcal{M}_{D}$:
{\small
\begin{align}
\label{loss:md}
\mathcal{M}_{D} &= \sum_{t=1}^{T}\mathbb{E}_{q(\mathbf u_s,a^t,\mathbf u_{ns})}\big[\log(D([\mathbf u_s,a^t,\mathbf u_{ns}]))\big]\nonumber\\
&+\mathbb{E}_{q(\mathbf u_s)q(a^t,\mathbf u_{ns})}\big[\log{(1-D([\mathbf u_s,a^t,\mathbf u_{ns}]))}\big]\nonumber\\&=\sum_{t=1}^{T}
\mathbb{E}_{q(\mathbf u_s,a^t,\mathbf u_{ns})}\big[\log(D([\mathbf u_s,a^t,\mathbf u_{ns}]))\big]\nonumber\\
&+\mathbb{E}_{q(\mathbf u_s,a^t,\mathbf u_{ns})}\big[\log{(1-D(perm[\mathbf u_s,a^t,\mathbf u_{ns}]))}\big],
\end{align}}
where $perm[\mathbf u_s,a^t,\mathbf u_{ns}]$ denotes the randomized alteration of the relative sequence between $(a^t,\mathbf u_{ns})$ and $\mathbf u_s$.

\begin{figure}[t]
\vspace*{-\baselineskip}
\begin{minipage}{\columnwidth}
\begin{algorithm}[H]
   \caption{Optimization procedure for \sysname{}}
   \label{alg:optimization1}
\begin{algorithmic}[1]
   \STATE {\bfseries Input:} sequential source labeled datasets $\mathcal{S}$ with $T$ domains; static feature extractor $E^{s}$, $E^{ns}$; dynamic inference networks $E^{v1}$, $E^{v2}$ and their corresponding prior networks (LSTM) $F^{v1}$, $F^{v2}$; decoder $D^{s}$, $D^{ns}$; discriminator $D$; classifier $C$.
   \STATE \textbf{Initialize} $E^{s},E^{ns},E^{v1},E^{v2},F^{v1},F^{v2},D^{s},$ $ D^{ns},D,C$
   \STATE \textbf{Assign} $\mathbf{u}^0_{v1}, \mathbf{u}^0_{v2} \gets \mathbf{0}$

   \FOR {$t = 1,2,...,T$}
   \STATE Generate prior distribution $p(\mathbf{u}^t_{v1}|\mathbf{u}^{<t}_{v1})$ via $F^{v1}$
  \STATE Generate prior distribution $p(\mathbf{u}^t_{v2}|\mathbf{u}^{<t}_{v2})$ via $F^{v2}$
      \FOR {$i = 1, 2, ...$}
      \STATE Sample a batch of data $(\mathbf{x}^t_s,\mathbf{x}^t_{ns},{a}^t, {y}^t)$ from $\mathcal{D}_t$
      \STATE Calculate $\mathcal{L}_{\sysname{}}$ by Eq.~\ref{eq:loss_min}
      \STATE Update $E^{s}$, $E^{ns}$,$E^{v1}$, $E^{v2}$,$F^{v1}$, $F^{v2}$
      ,$D^{s}$, $D^{ns}$ and $C$ by $\mathcal{L}_{\sysname{}}$
      \STATE Calculate $\mathcal{M}_D$ by Eq.~\ref{loss:md} 
      \STATE Update $D$ by $\mathcal{M}_D$
      \ENDFOR
   \ENDFOR
\end{algorithmic}
\end{algorithm}
\end{minipage}

\end{figure}
\subsection{Ultimate Objective Function}
We denote all parameters of \sysname{}, including all encoders, decoders, and prior networks (LSTMs), as $\theta$, and the parameters of discriminator $D$ as $\psi$.
Summing up the preceding sections, the training objectives of the model can be summarized into two phases as follows:
{\small
\begin{align}
\ & min_{\theta}{\ \mathcal{L}_{\sysname{}}} := -{\text{ELBO}} + \lambda_{f} \mathcal{L}_{f}+ \lambda_{tc} \mathcal{L}_{TC},
\label{eq:loss_min}
\\
\ & max_{\psi}{\ \mathcal{M}_{D}}.
\label{eq:loss_max}
\end{align}
}

After the completion of training within the \sysname{} framework (Algorithm.~\ref{alg:optimization1}), we require the trained static feature extractor $E^{s}$ and $E^{ns}$ to obtain semantic information ($u_s$ and $u_{ns}$). Finally, the classifier $C$ is utilized for prediction by inputting both $u_s$ and $u_{ns}$ alongside sensitive attribute $a$.

\section{Theoretical Guarantee of \sysname{}}
Due to the usual representation of ELBO as a sum of multiple terms, we delve into its equivalent optimization objective in theoretical analysis.
\begin{lemma}
\label{lemma1}
In the vanilla VAE, the KL divergence $ \text{KL}(q(\mathbf u|\mathbf x)||p(\mathbf u|\mathbf x))$ can be represented as
{\small
\begin{align}
\text{KL}(q(\mathbf u|\mathbf x)||p(\mathbf u))-E_{q(\mathbf u|\mathbf x)}[\log p(\mathbf x|\mathbf u)]+\log p(\mathbf x).
\end{align}}
\end{lemma}
Based on Lemma 1, we can derive the Evidence Lower Bound (ELBO) of the vanilla VAE in the following formula: \begin{align}
\text{ELBO} = \log p(\mathbf x)-\text{KL}(q(\mathbf u|\mathbf x)||p(\mathbf u|\mathbf x))
\end{align}
It means that optimizing the ELBO of VAEs is equivalent to optimizing $ \text{KL}(q(\mathbf u|\mathbf x)||p(\mathbf u|\mathbf x))$. We denote the samples from the source domains as $X_s^{1:T}$ and $X_{ns}^{1:T}$, while the features of samples from the unseen target domains are represented as $X_s^{T+m}$ and $X_{ns}^{T+m}$ for $m\ge 1$. The relationship between the source domains and the target domains can be expressed as follows. 
\begin{theorem}
\label{sec:th1}
 The KL divergence between $q(\mathbf{u}_s,\mathbf{u}_{ns}|\mathbf x_s^{T+m},a^{T+m},\mathbf x_{ns}^{T+m})$ and the unknown domain-invariant ground truth distribution $p(\mathbf{u}_s,\mathbf{u}_{ns}|\mathbf x_s^{T+m},a^{T+m},,\mathbf x_{ns}^{T+m})$ can be bounded as follows:
 {\small
\begin{align}
     &\text{KL}(q(\mathbf{u}_s,\mathbf{u}_{ns}|\mathbf x^{T+m},a^{T+m})||p(\mathbf{u}_s,\mathbf{u}_{ns}|\mathbf x^{T+m},a^{T+m})  \nonumber\\ 
     \leq&\inf_{I\in \mathcal{I}} [\sum_{i\in I} \beta_i (\text{KL}(q(\mathbf{u}_s|\mathbf x_s^{1:T,i},a^{1:T,i})||p(\mathbf{u}_s|\mathbf x_{s}^{1:T,i}))\nonumber\\&+\text{KL}(q(\mathbf{u}_{ns}|\mathbf x_{ns}^{1:T,i},a^{1:T,i})||p(\mathbf{u}_{ns}|\mathbf x_{ns}^{1:T,i})))],\nonumber
\end{align}}
where $\mathbf x_s^{1:T,i},a^{1:T,i}$ and $\mathbf x_{ns}^{1:T,i}$ denotes features with index $i$ in source domains. The feasible set $\mathcal I$~\cite{wang2021variational} and constant $\beta_i$ are defined in Appendix A.3. Semantic information $\mathbf{u}_s$ and $\mathbf{u}_{ns}$ are defined in Section~\ref{sec:Causal Structure}.
\end{theorem}

This inequality expresses that the ELBO on the target domains can be optimized by separately optimizing the ELBO concerning $X_s$ and $X_{ns}$ on the source domains. 
Therefore, Theorem~\ref{sec:th1} ensures that \sysname{}~is a rational and effective methodology. The detailed proof of Theorem~\ref{sec:th1} is provided in Appendix A.4.

\section{Experiments}
    \label{sec:exp}
    \subsection{Datasets}

\textbf{FairCircle} is a synthetic dataset containing 12 domains. 
For each domain, followed by \cite{zafar2017fairness}, we generate 2000 binary class labels uniformly at random and assign a two-dimensional feature vector $\mathbf{x}=[x_s,x_{ns}]^T$ per label by sampling from two distinct Gaussian distributions: $\mathbb{P}(\mathbf{x}|y=0)=\mathcal{N}(\mathbf{\mu_{0}},[10,1;1,3])$ and $\mathbb{P}(\mathbf{x}|y=1)=\mathcal{N}(\mathbf{\mu_{1}},[5,1;1,5])$, where $\mathbf{\mu_{0}}$ and $\mathbf{\mu_{1}}$ will changed by domain.
Sensitive attributes of data samples are drawn from a Bernoulli distribution $\mathbb{P}(a=1)=\frac{\mathbb{P}(\mathbf{x'}|y=1)}{\mathbb{P}(\mathbf{x'}|y=1)+\mathbb{P}(\mathbf{x'}|y=0)}$, where $\mathbf{x'}=[\cos(\phi),-\sin(\phi);\sin(\phi),\cos(\phi)][{x}_s;1]$ is simply a rotated vector related to $x_s$. 
The $\phi$ controls the correlation between the sensitive attribute and the class labels. 
The $\phi$ in each domain is a random number between $\frac{\pi}{8}$ and $\frac{\pi}{4}$.
The closer $\phi$ is to zero, the higher the correlation. To construct multiple sequentially changing domains, we uniformly sampled 12 values of $\mu_0$ and $\mu_1$ from two circular arcs with radii of 25 and 34, respectively, to simulate the variation in data distribution. The visualization of the dataset is provided in Appendix B.1.

\begin{table*}[!h]
    \scriptsize
    \centering
    \renewcommand{\tabcolsep}{1.5mm}

    \begin{tabular}{l|c c | c c c c c c | c c c c c c}
        \toprule
        & \multicolumn{2}{c|}{\textbf{FairCircle}} & \multicolumn{6}{c|}{\textbf{Adult}} & \multicolumn{6}{c}{\textbf{Chicago Crime}} \\
        \cmidrule(lr){1-1}\cmidrule(lr){2-3} \cmidrule(lr){4-9} \cmidrule(lr){10-15}
        \multirow{2}{*}{Methods}
        & \multirow{2}{*}{Acc $\uparrow$}
        & \multirow{2}{*}{\makecell[c]{TCE $\downarrow$ \\($\times 10$)}}
        & \multirow{2}{*}{Acc $\uparrow$}
        & \multirow{2}{*}{\makecell[c]{TCE $\downarrow$ \\($\times 10$)}}
        & \multicolumn{4}{c|}{CE $\downarrow$ ($\times 10$)}
        & \multirow{2}{*}{Acc $\uparrow$}
        & \multirow{2}{*}{\makecell[c]{TCE $\downarrow$ \\($\times 10$)}}
        & \multicolumn{4}{c}{CE $\downarrow$ ($\times 10$)} \\
        \cmidrule(lr){6-9} \cmidrule(lr){12-15}
        & & & & & $o_{00}$ & $o_{01}$ & $o_{10}$ & $o_{11}$
        & & &  $o_{00}$ & $o_{01}$ & $o_{10}$ & $o_{11}$ \\
        \midrule
        DIVA~\cite{ilse2020diva} & 69.10   & 1.15
        & \underline{68.04} & 0.81 & 0.88 & 0.62 & \underline{0.34} & 0.86
        & \textbf{56.19} & {1.68} & 1.68 & {1.46} & {1.84} & 1.75 \\
        
        LSSAE~\cite{qin2022generalizing} & \textbf{89.25} & 5.03
        & 57.79 & 1.91 & 2.96 & 3.64 & 1.70 & 1.67
        & 53.72 & 0.85 & 0.77 & 0.93 & 0.90 & 0.77 \\
        
        MMD-LSAE~\cite{qin2023evolving} & 82.79   & 0.70
        & 60.34 & 1.60 & 1.17 & 1.35& 1.05 & 1.68
        & 53.83 & \underline{0.35} & \underline{0.23} & \underline{0.41 }&\underline{0.36} & \underline{0.31}\\
        \midrule
        CVAE~\cite{sohn2015learning} & 49.99   & 0.18
        & 61.83 & 0.56 & 0.53 & 0.55 & 0.51 & 0.57
        & 54.43 & {0.72} & 0.67 & {0.70} & {0.74} & 0.77 \\
        
        CEVAE~\cite{louizos2017causal} & 49.99 & 0.34
        & 62.49 & 0.69 & 0.68 & 0.69 & 0.69 & 0.69
        & 54.23 & 0.42 & 0.40 & 0.43 & 0.42 & 0.44 \\
        
        mCEVAE~\cite{pfohl2019counterfactual} & 63.30   & 0.28
        & 61.05 & 0.48 & 0.45 & \underline{0.35} & 0.50 & 0.48
        & {51.83} & \textbf{0.01} & \textbf{0.01} & \textbf{0.01} & \textbf{0.01} & \textbf{0.01} \\
        
        DCEVAE~\cite{kim2021counterfactual} & 53.25   & \underline{0.18}
        & {62.69} & \underline{0.39} & \underline{0.39} & 0.38 & {0.39} & \underline{0.38}
        & 51.29 & 0.44 & 0.48 & 0.45 & 0.44 & 0.39 \\
        \midrule
        \sysname{} (Ours) & \underline{88.70} & \textbf{0.12}
        & \textbf{69.85} & \textbf{0.22} & \textbf{0.10} & \textbf{0.01} & \textbf{0.17} & \textbf{0.26}
        & \underline{55.93} & \textbf{0.01} & \textbf{0.01} & \textbf{0.01} & \textbf{0.01} & \textbf{0.01} \\
        
        \bottomrule
    \end{tabular}
        \caption{Accuracy outcomes and TCE value results across the three datasets. Within the experiment, the variable $\mathbf{O}$ comprises two attributes, where $o_{ij}$ denotes the first attribute as $i$ and the second attribute as $j$.}
    \label{table:total causal effect and counterfactual effect}
\end{table*}

\textbf{Adult}~\cite{kohavi1996scaling} contains a diverse set of attributes pertaining to individuals in the United States. The dataset is often utilized to predict whether an individual's annual income exceeds 50,000 dollars, making it a popular choice for binary classification tasks. We categorize gender as a sensitive attribute. Income is designated as the dependent variable $Y$. Race, age, and country of origin constitute the set $X_{ns}$, while the remaining variables comprise the set $X_s$~\cite{zhao2021you,grari2021fairness,kim2021counterfactual}. We divided the samples into 18 domains based on age, ranging from younger to older. Specifically, the source domain tends to represent a younger demographic, while the target domain tends to represent an older demographic.

\textbf{Chicago Crime}~\cite{zhao2020unfairness} dataset includes a comprehensive compilation of criminal incidents in different communities across Chicago city in 2015. 
We use race (\textit{i.e.,} black and non-black) as the sensitive attribute. To better delineate between $X_s$ and $X_{ns}$, we measured the Pearson Product-Moment Correlation Coefficients (PPMCC) values between each feature and sensitive attribute (Appendix B.2). This was done to gauge their correlation and aid in the partitioning process. Grocery count, per capita income, aged 25+ without high school diploma, and housing crowd of origin constitute the set $X_{ns}$, while the remaining variables comprise the set $X_s$. The dataset was collected over time, and as a result, we partition the data into 18 domains based on chronological order. The target domain consists of the most recent samples.

\subsection{Baseline Methods}
We evaluate the proposed \sysname{} against seven baseline methods. These baselines are selected from two perspectives:  approaches that utilize causal structures to tackle evolving domain generalization (DIVA~\cite{ilse2020diva}, LSSAE~\cite{qin2022generalizing}, and MMD-LSAE~\cite{qin2023evolving}), and methods that utilize causal structures to address counterfactual fairness (CVAE~\cite{sohn2015learning}, CEVAE~\cite{louizos2017causal}, mCEVAE~\cite{pfohl2019counterfactual}, and DCEVAE~\cite{kim2021counterfactual}).


\subsection{Evaluation Metrics}
We employed two metrics, total causal effect and counterfactual effect, to evaluate the fair classification. Assuming $A$ is the intervention target of the do-operator, $Y$ is influenced by this intervention. The post-intervention distribution of $Y$ mentioned in Section~\ref{sec:scm} can be further abbreviated as $\mathbb P(y_a)$. 
\begin{Definition}[Total Causal Effect (TCE) \cite{pearl2009causality}]
    The total causal effect of the value change of $A$ from $a$ to $\neg a$ on $Y=y$ is given by $TCE(a, \neg a)= | \mathbb P(y_{a})-\mathbb P(y_{\neg a})|$.
\end{Definition}
\begin{Definition}[Counterfactual Effect (CE) \cite{shpitser2008complete}]
    Given context $O=o$, the counterfactual effect of the value change of $A$ from $a$ to $\neg a$ on $Y=y$ is given by $CE(a,\neg a|\mathbf{o})=|\mathbb P(y_{a}|\mathbf{o})-\mathbb P(y_{\neg a}|\mathbf o)|$.
\end{Definition}
Smaller TCE and CE indicate that the prediction results are more stable in the counterfactual generation of changing the sensitive attribute, implying greater fairness~\cite{wu2019pc}. For the Adult dataset, we set context  of counterfactual effect as $O=\{$race, native country$\}$. For the Crime dataset, we set context of counterfactual effect as $O=\{$grocery count, per capital income$\}$. In both two datasets, $o_{ij}$ denotes the first attribute as $i$ and the second attribute as $j$.

\subsection{Experimental Setup}
We partitioned the domains into source, intermediary, and target domains by the ratio $(\frac{1}{2}:\frac{1}{6}:\frac{1}{3})$. The source domains are employed for training the \sysname{}, while the intermediary domains serves as the validation set. All evaluations are conducted within the target domains. For the FairCircle dataset, direct computation of its counterfactual effect (CE) is unfeasible because its features are randomly sampled continuous numerical values. As for the other two datasets, both the total causal effect (TCE) and CE were employed for evaluation purposes.
For all the encoders, decoders, classifiers, and discriminators, we employed the most common fully connected layers and ReLU activation functions. The specific architecture details can be found in Appendix B.3.

\subsection{Results Analysis}

\begin{figure*}
\label{fig:curve1}
  \centering

\begin{subfigure}[b]{1.5\columnwidth}
    \centering
    \includegraphics[width=\textwidth]{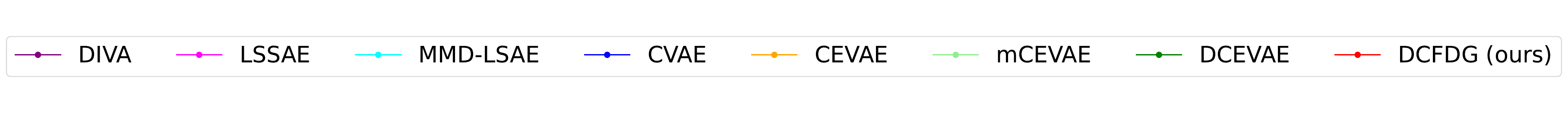}
\end{subfigure}%
\\
\begin{subfigure}[b]{0.1666\textwidth}
    \centering
    \includegraphics[width=\textwidth]{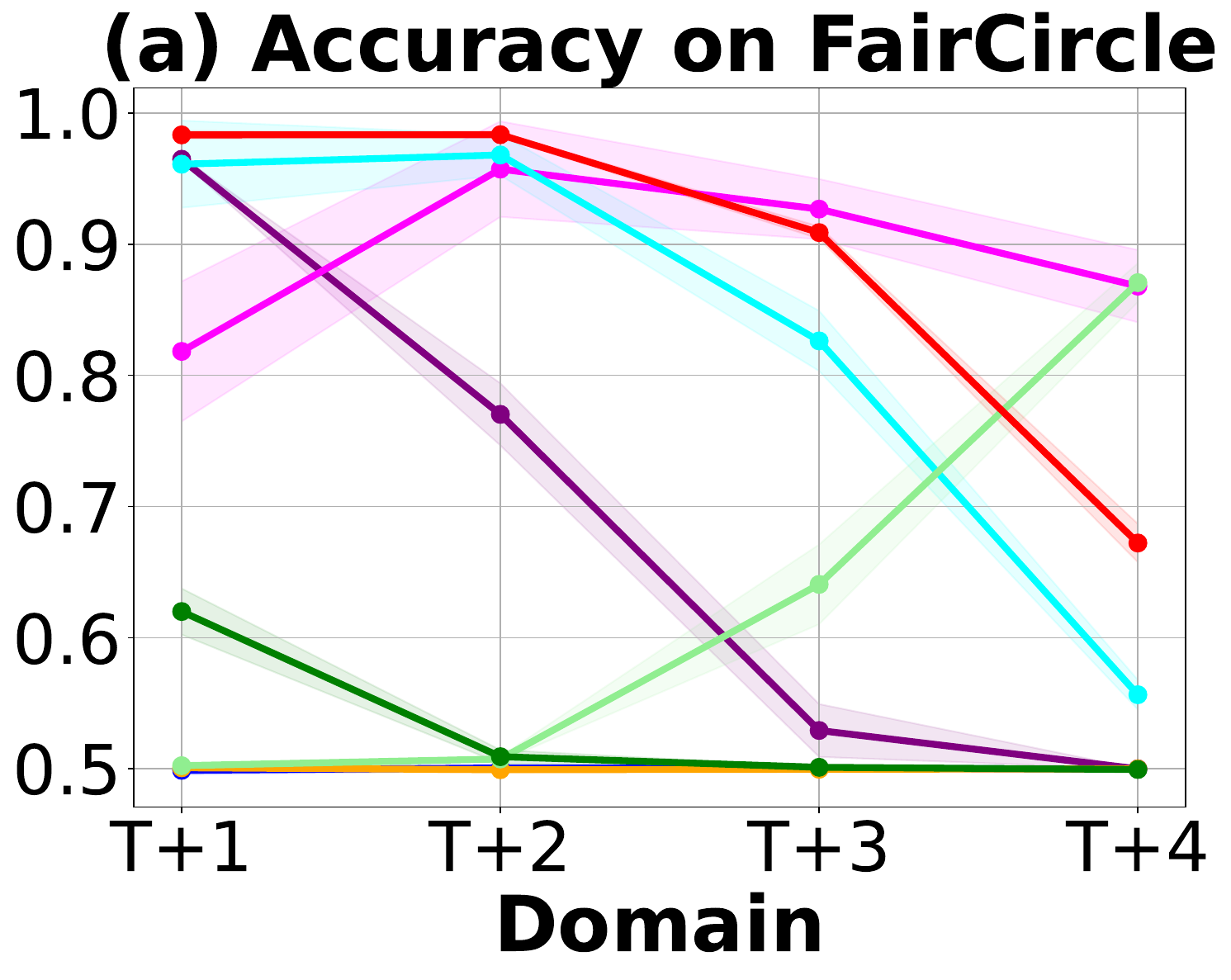}
\end{subfigure}%
\begin{subfigure}[b]{0.1666\textwidth}
    \centering
    \includegraphics[width=\textwidth]{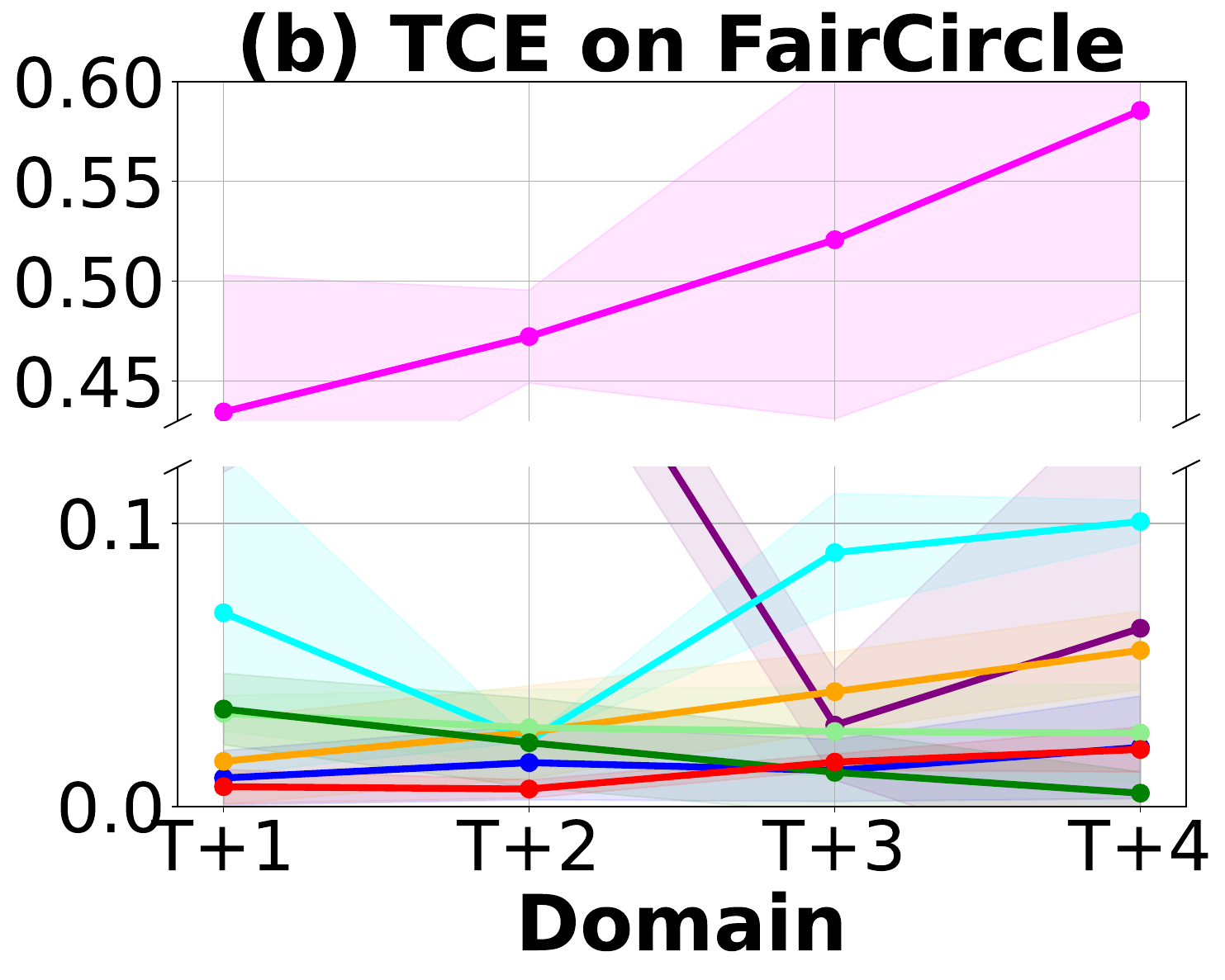}
\end{subfigure}%
\begin{subfigure}[b]{0.1666\textwidth}
    \centering
    \includegraphics[width=\textwidth]{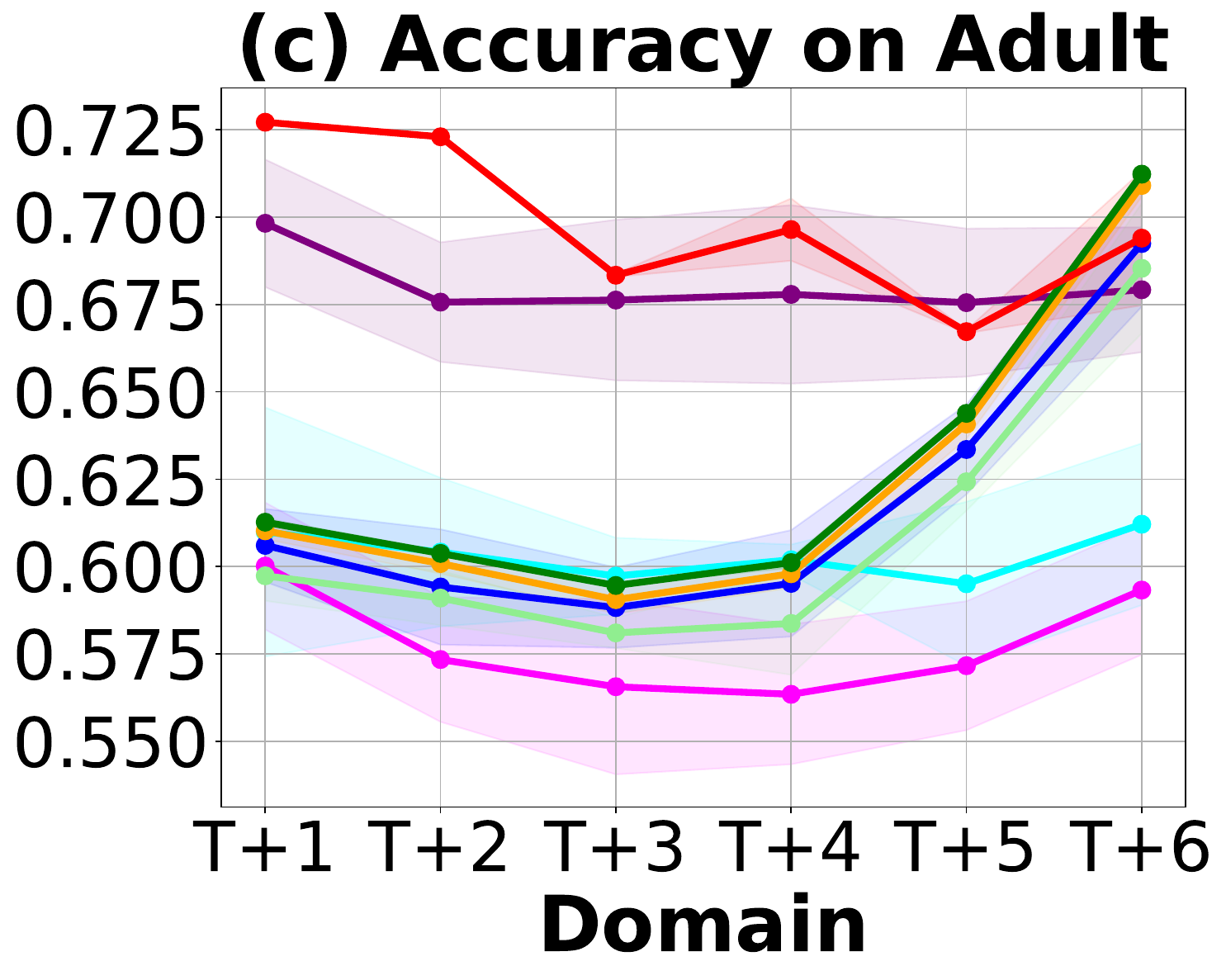}
\end{subfigure}%
\begin{subfigure}[b]{0.1666\textwidth}
    \centering
    \includegraphics[width=\textwidth]{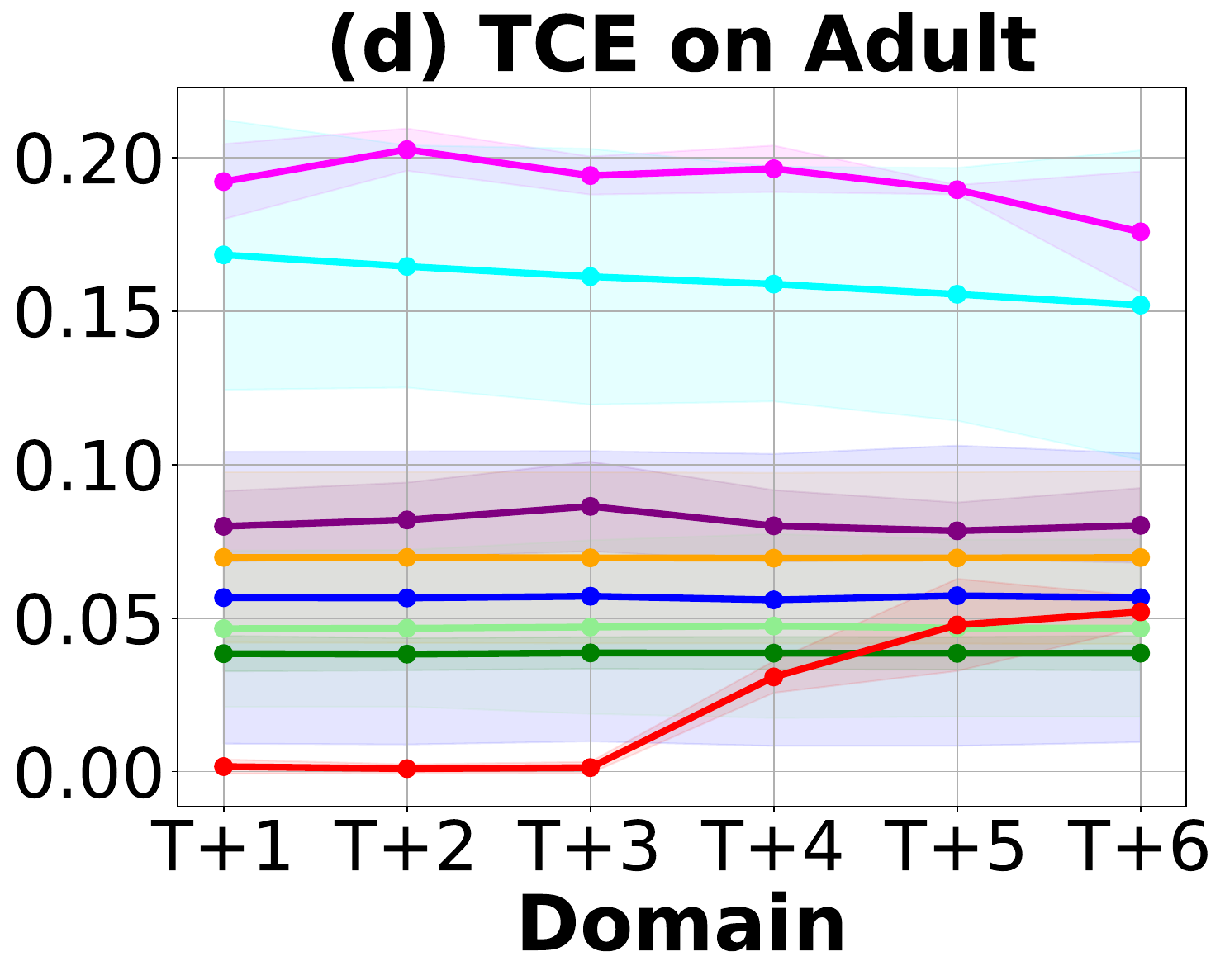}
\end{subfigure}%
\begin{subfigure}[b]{0.1666\textwidth}
    \centering
    \includegraphics[width=\textwidth]{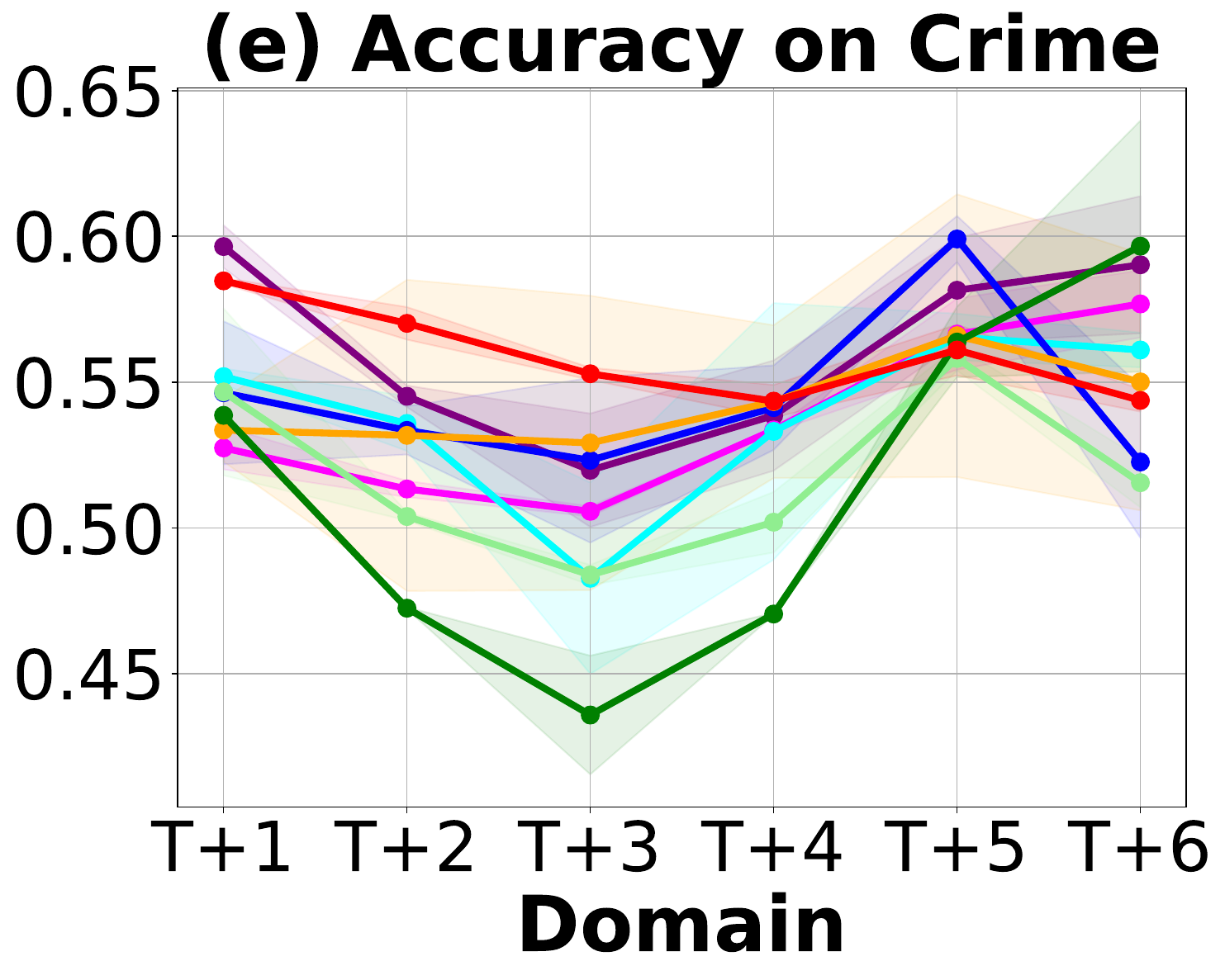}
\end{subfigure}%
\begin{subfigure}[b]{0.1666\textwidth}
    \centering
    \includegraphics[width=\textwidth]{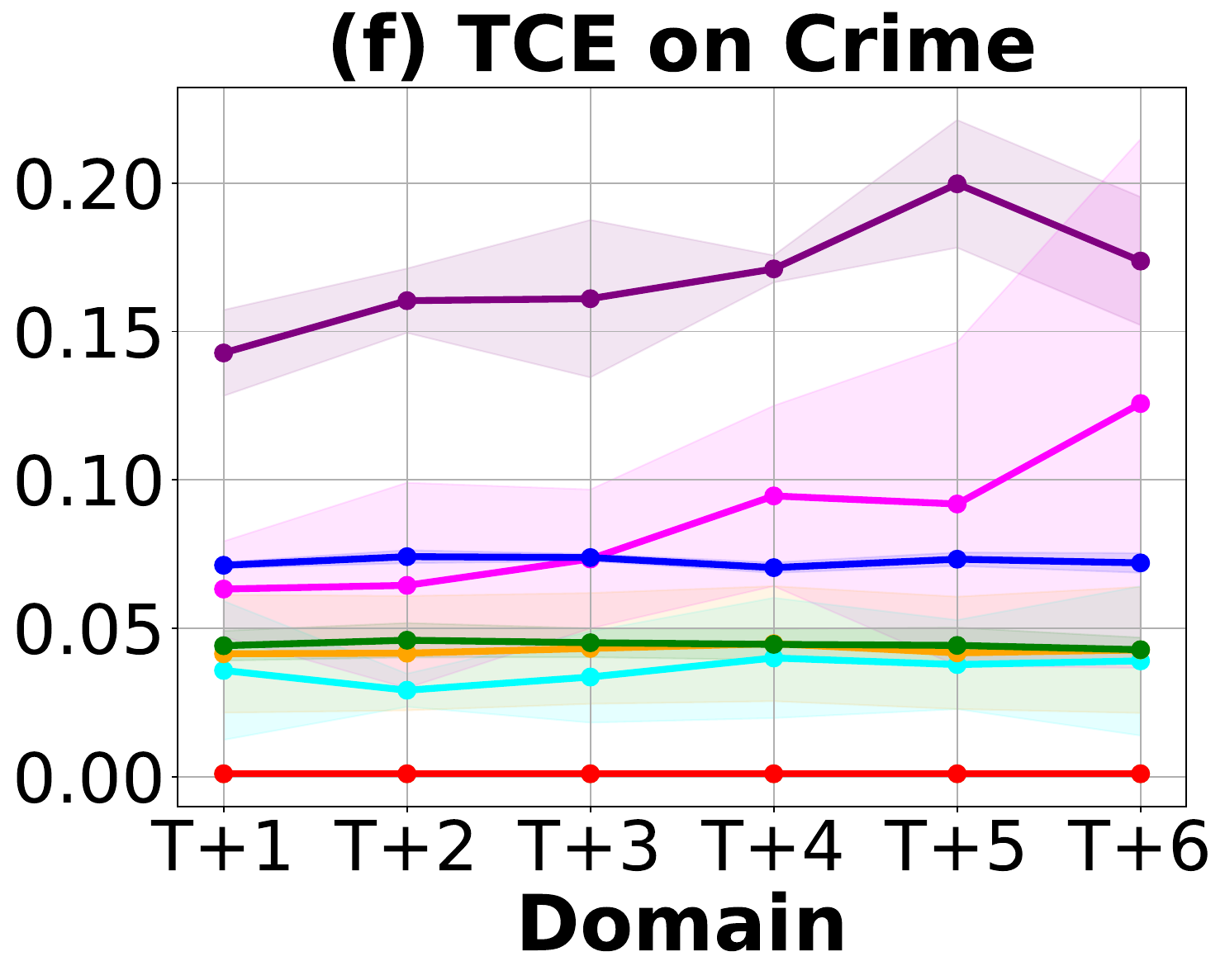}
\end{subfigure}%
  \caption{Accuracy and total causal effect for each testing domain. The 1st, 3rd, and 5th figures illustrate the accuracy curves, while the 2nd, 4th, and 6th figures depict the total causal effect curves.}
\vspace{-3mm}
\end{figure*}

\textbf{Overall Performance.}
We computed the mean performance across all testing domains, as depicted in Table~\ref{table:total causal effect and counterfactual effect}. Smaller values of TCE and CE indicate closer adherence of the classification outcomes to counterfactual fairness. To facilitate observation, the reported results encapsulate the values of TCE and CE across all outcomes. Across the three datasets, \sysname{} consistently demonstrates favorable generalization capabilities to unknown domains compared to other approaches, achieving optimal performance. Notably, its pronounced superiority in accuracy on the FairCircle dataset is believed to stem from the discernible advantage exhibited as the data distribution between each domain varies to a greater extent. Regarding TCE and CE, \sysname{} consistently achieves optimal or near-optimal outcomes. This underscores the resilience of our approach to maintaining high performance while simultaneously upholding fairness principles. For the Chicago Crime dataset, while there hasn't been a substantial improvement in decision accuracy, it is noteworthy that both its TCE and CE values are considerably lower than the highest accuracy method: DIVA. In other words, in the context of comparable accuracy levels, fairness significantly outperforms alternative methods.

\textbf{Performance Across Each Domain.}
In Figure 3, we present the results across each testing domain. For the FairCircle dataset, there are four testing domains, while the Adult and Chicago Crime datasets have six testing domains each. The 1st, 3rd, and 5th figures represent accuracy outcomes, with higher curves indicating superior performance. The 2nd, 4th, and 6th figures illustrate TCE results, with lower curves signifying enhanced compliance with counterfactual fairness, concurrently denoted by the shaded regions representing standard deviations. Across all testing domains, \sysname{} consistently maintains superior accuracy and minimal TCE values. 
 Regarding the tabulated data encompassing the mean and standard deviation of all three metrics across each domain, we present this information uniformly within the Appendix B.6.

\subsection{Ablation Study}
\begin{table}[!t]
    \scriptsize
    \centering
    \renewcommand{\tabcolsep}{1mm}

    \begin{tabular}{l | c c | c c }
        \toprule
         
         & \multicolumn{2}{c|}{\textbf{Adult}} & \multicolumn{2}{c}{\textbf{Chicago Crime}} \\
\cmidrule(lr){1-1}
        \cmidrule(lr){2-3} \cmidrule(lr){4-5} 
        \multirow{2}{*}{Metric}
        & \multirow{2}{*}{Acc $\uparrow$}
        & \multirow{2}{*}{\makecell[c]{TCE $\downarrow$\\($\times 10$)}}
        & \multirow{2}{*}{Acc $\uparrow$}
        & \multirow{2}{*}{\makecell[c]{TCE $\downarrow$\\($\times 10$)}}
        \\
        \\
        \midrule
        w/o disentanglement
        & {71.48} & {0.47}
        & {54.43} &{1.61} \\
        
        w/o fairness loss 
        & \textbf{72.24} & {2.76} 
        & {54.89} & 1.75 \\
    
        \sysname{} 
        & {69.85} & \textbf{0.22} 
        & \textbf{55.93} & \textbf{0.01} \\
        
        \bottomrule
    \end{tabular}
    
            \caption{Ablation study results across the two datasets. 
The results in the table represent the mean values of all test domain outcomes.}

    \label{table:ablation}
\end{table}
We evaluate the effect of components in the design of \sysname{}'s objective. We have specifically examined two variants of \sysname{} as follows.

\textbf{Without Disentanglement.} We attempted to refrain from decoupling features into domain-specific and semantic information, opting instead for utilizing a globally modeled dynamic Gaussian distribution for predictions. As indicated in Table~\ref{table:ablation}, the absence of feature decoupling adversely impacted classification fairness, particularly evident in the Crime dataset.

\textbf{Without Fairness Loss.} We eliminated the loss associated with counterfactual fairness to assess changes in the outcomes. Despite achieving a marginal advantage in prediction accuracy on the adult dataset, a sharp increase in the TCE value resulted in unfair classification outcomes (Table~\ref{table:ablation}).

Experimental results regarding the CE values can be found in Appendix B.4. The above experiments indicate that decoupling domain-specific information and incorporating the fairness loss are both indispensable for ensuring counterfactual fairness. 

\subsection{Fairness-accuracy Trade-off}
\begin{figure}[!t]
  \centering
  
 \begin{subfigure}[b]{0.45\columnwidth}
    \centering
    \includegraphics[width=\textwidth]{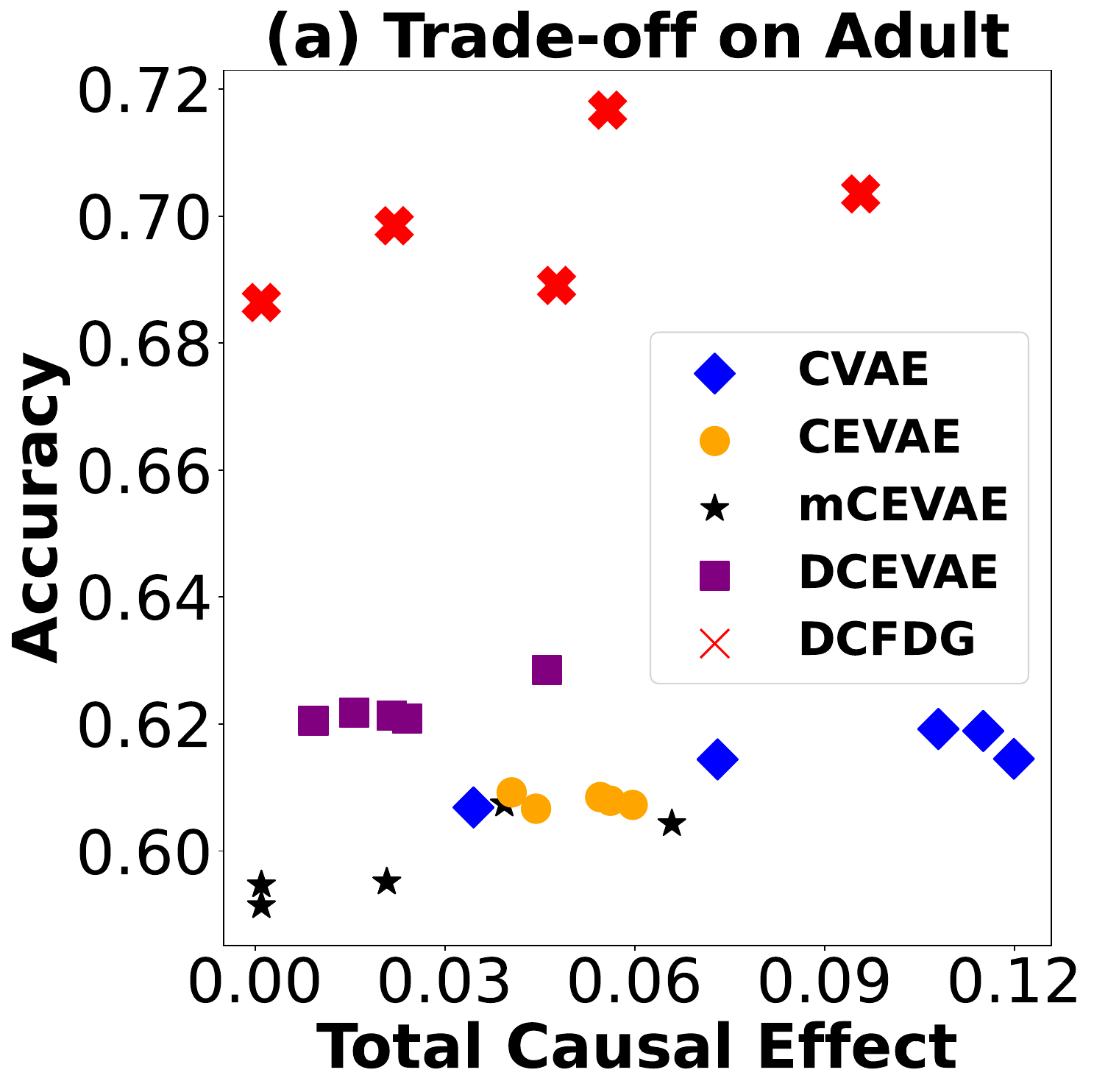}
    \label{fig:faircircle_acc}
  \end{subfigure}%
  \begin{subfigure}[b]{0.45\columnwidth}
    \centering
    \includegraphics[width=\textwidth]{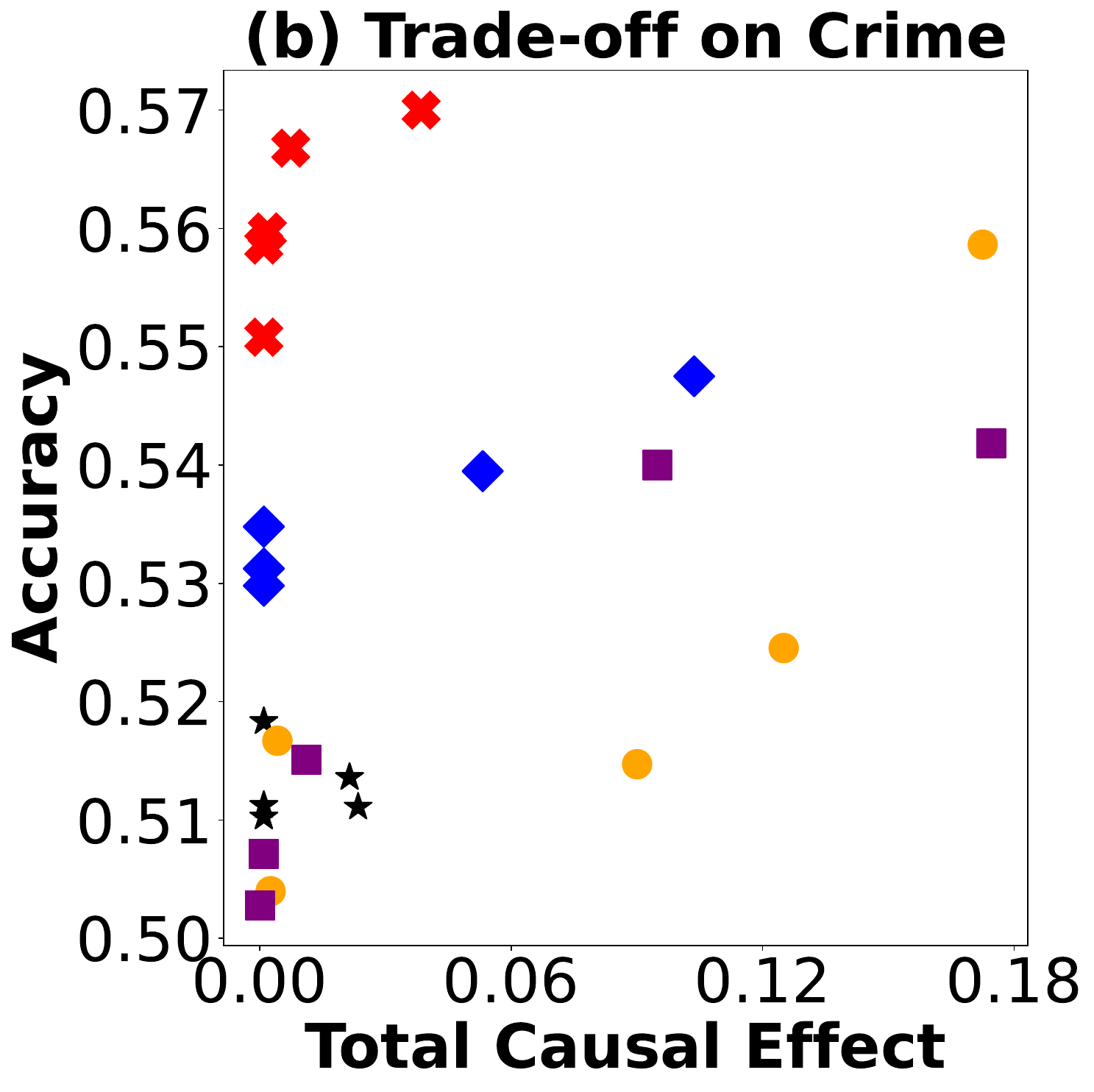}
    \label{fig:faircircle_te}
  \end{subfigure}%
  \caption{Fairness-accuracy Trade-off on Adult and Crime. Each baseline is represented by five data points, corresponding to the outcomes under five distinct fairness parameter $\lambda_f$.}
  \label{img:trade-off}
\end{figure}
Due to the absence of fairness loss in certain baselines, we compare our method with four baselines about the trade-off between accuracy and fairness on target domains under different parameters. We varied the parameter $\lambda_f$ across five values ([0.02, 0.1, 0.2, 0.5, 1]) to obtain the results of each baseline under these five settings. In Figure~\ref{img:trade-off}, the horizontal axis represents TCE values, and the vertical axis represents accuracy, indicating that data points tending towards the upper-left corner exhibit superior performance. Experimental results regarding the CE values can be found in Appendix B.5. All the results demonstrate that \sysname{} achieves the best overall performance.

\section{Conclusion}
    \label{sec:conclusion}
     In summary, this paper has proposed a novel framework, \sysname{}, to address issues of fairness within continuously evolving dynamic environments. This method disentangles exogenous variables based on the relationships among sensitive attributes, domain-specific information, and semantic information, partitioning them into four latent variables. 
By leveraging these latent variables, a causal structure is constructed for our method. We establish an appropriate model and optimize the corresponding objective function through this causal graph.
Theoretical analysis and experimental validation attest to the efficacy of \sysname{}.

\section*{Acknowledgements}
This work is supported by the National Natural Science Foundation of China program (NSFC \#62272338).

\bibliographystyle{named}
\bibliography{refer}

 \appendix
 \clearpage
 \onecolumn

 \section{Appendix}
 
\subsection{Introduction}
This is the supplementary material for the paper `Towards Counterfactual Fairness-aware Domain Generalization in Changing Environments'. 

\subsection{Notations}
\label{notaions}
\begin{table*}[!h]
\caption{Important notations and their description.}
\begin{center}

\begin{tabular}{l|l}
\toprule
\textbf{Notation} & \textbf{Description}\\
\midrule
$T$ & Total number of training domains\\
$t$ & Indices of domains\\
$\mathcal{D}_t$ & Domain at time $t$\\
$X_s$ & Features caused by sensitive attribute\\
$X_{ns}$ & Features not caused by sensitive attribute\\
$A$ & Sensitive attribute\\
$Y$ & Ground truth of samples\\
$U_s$ & Semantic information caused by sensitive attribute\\
$U_{ns}$ & Semantic information not caused by sensitive attribute\\
$U_{v1}$ & Domain specific information caused by sensitive attribute\\
$U_{v2}$ & Domain specific information not caused by sensitive attribute\\
$E^s$ & Encoder for encoding $U_s$\\
$E^{ns}$ & Encoder for encoding $U_{ns}$\\
$E^{v1}$ & Encoder for encoding $U_{v1}$\\
$E^{v2}$ & Encoder for encoding $U_{v2}$\\
$D^s$ & Decoder for decoding $X_s$\\
$D^{ns}$ & Decoder for decoding $X_{ns}$\\
$C$ & Classifier for predicting $\hat Y$\\
\bottomrule
\end{tabular}

\label{tab.notations}
\end{center}
\end{table*}

\subsection{Theoretical Guarantee of DCFDG}
\label{Appendix:theory}
\begin{lemma}
\label{lemma1}
In the vanilla VAE, the KL divergence $ \text{KL}(q(\mathbf u|\mathbf x)||p(\mathbf u|\mathbf x))$ can be represented as
\begin{align}
\text{KL}(q(\mathbf u|\mathbf x)||p(\mathbf u))-E_{q(\mathbf u_c|\mathbf x)}[\log p(\mathbf x|\mathbf u_c)]+\log p(\mathbf x).
\end{align}
\end{lemma}
Based on Lemma 1, we can derive the Evidence Lower Bound (ELBO) of the vanilla VAE in the following fomula: \begin{align}
\text{ELBO} = \log p(\mathbf x)-\text{KL}(q(\mathbf u|\mathbf x)||p(\mathbf u|\mathbf x))
\end{align}
It means that optimizing the ELBO of VAEs is equivalent to optimizing $ \text{KL}(q(\mathbf u|\mathbf x)||p(\mathbf u|\mathbf x))$. We denote the samples from the training domain as $X_s^t$ and $X_{ns}^t$ for $t \in \{1,2,...,T\}$, while the features of samples from the unseen testing domain are represented as $x_s^{T+m}$ and $X_{ns}^{T+m}$ for $m\ge 1$. And all the training data can be represented as $X_s^{1:T}$ and $X_{ns}^{1:T}$.
\begin{definition}
\label{def1}
    Based on the previous work~\cite{wang2021variational}, 
we will consider scenarios involving thge sensitive attribute $A$ and the partitioning of $X^t$ into $X_s^t$ and $X_{ns}^t$. There exists a non-empty feasible set $\mathcal{I}$ which is defined as
\begin{align}
    \mathcal{I} = &\{I | q(\mathbf{u}_s,\mathbf{u}_{ns}|\mathbf x_s^{T+m},a^{T+m},\mathbf x_{ns}^{T+m}) \leq \sum_{i\in I} \beta_i q(\mathbf{u}_s,\mathbf{u}_{ns}|\mathbf x_s^{1:T,i},a^{1:T,i},\mathbf x_{ns}^{1:T,i})\} \nonumber\\ &\cap \{I | \phi_c(\mathbf x_s^{T+m},a^{T+m},\mathbf x_{ns}^{T+m}) = \phi_c(\mathbf x_s^{1:T,i},a^{1:T,i},\mathbf x_{ns}^{1:T,i}) , 
\end{align}
where $I$ is the index set, and $\phi_c$ is a function to extract features' semantic information.
\end{definition}
\begin{theorem}
\label{th1}
 The KL divergence between $q(\mathbf{u}_s,\mathbf{u}_{ns}|\mathbf x_s^{T+m},a^{T+m},\mathbf x_{ns}^{T+m})$ and the unknown domain-invariant ground truth distribution $p(\mathbf{u}_s,\mathbf{u}_{ns}|\mathbf x_s^{T+m},a^{T+m},\mathbf x_{ns}^{T+m})$ can be bounded as follows:
\begin{align}
     &\text{KL}(q(\mathbf{u}_s,\mathbf{u}_{ns}|\mathbf x_s^{T+m},a^{T+m},\mathbf x_{ns}^{T+m})||p(\mathbf{u}_s,\mathbf{u}_{ns}|\mathbf x_s^{T+m},a^{T+m},\mathbf x_{ns}^{T+m}))  \nonumber\\ 
     \leq&\inf_{I\in \mathcal{I}} [\sum_{i\in I} \beta_i (\text{KL}(q(\mathbf{u}_s|\mathbf x_s^{1:T,i},a^{1:T,i})||p(\mathbf{u}_s|\mathbf x_{s}^{T,i}))+\text{KL}(q(\mathbf{u}_{ns}|\mathbf x_{ns}^{1:T,i})||p(\mathbf{u}_{ns}|\mathbf x_{ns}^{1:T,i})))], 
\end{align}
where $\mathbf x_s^{1:T,i},a^{1:T,i}$ and $\mathbf x_{ns}^{1:T,i}$ denotes features with index $i$ in source domains. The feasible set $\mathcal I$~\cite{wang2021variational} is defined in Definition~\ref{def1}.

\end{theorem}

This inequality expresses that the ELBO on the target domains can be optimized by separately optimizing the ELBO concerning $X_s$ and $X_{ns}$ on the source domains. 
Therefore, Theorem~\ref{th1} ensures that DCFDG is a rational and effective methodology.

\subsection{Proof for Theorem~\ref{th1}}
\label{appendix:proof for th1}
$\forall  I \in \mathcal{I}$,we have
\begin{align}
     &\text{KL}(q(\mathbf{u}_s,\mathbf{u}_{ns}|\mathbf x_s^{T+m},a^{T+m},\mathbf x_{ns}^{T+m})||p(\mathbf{u}_s,\mathbf{u}_{ns}|\mathbf x_s^{T+m},a^{T+m},\mathbf x_{ns}^{T+m}))\nonumber\\=&\sum_{u_s}\sum_{u_{ns}}q(\mathbf{u}_s,\mathbf{u}_{ns}|\mathbf x_s^{T+m},a^{T+m},\mathbf x_{ns}^{T+m})log\frac{q(\mathbf{u}_s,\mathbf{u}_{ns}|\mathbf x_s^{T+m},a^{T+m},\mathbf x_{ns}^{T+m})}{p(\mathbf{u}_s,\mathbf{u}_{ns}|\mathbf x_s^{T+m},a^{T+m},\mathbf x_{ns}^{T+m})} \nonumber\\
     \le& \sum_{i\in I}\sum_{u_s}\sum_{u_{ns}}\beta_i q(\mathbf{u}_s,\mathbf{u}_{ns}|\mathbf x_s^{1:T,i},a^{1:T,i},\mathbf x_{ns}^{1:T,i})log\frac{q(\mathbf{u}_s,\mathbf{u}_{ns}|\mathbf x_s^{1:T,i},a^{1:T,i},\mathbf x_{ns}^{1:T,i})}{p(\mathbf{u}_s,\mathbf{u}_{ns}|\mathbf x_s^{1:T,i},a^{1:T,i},\mathbf x_{ns}^{1:T,i})} \nonumber\\
     =&\sum_{i\in I}\sum_{u_s}\sum_{u_{ns}}\beta_i q(\mathbf{u}_s|\mathbf x_s^{1:T,i},a^{1:T,i})q(\mathbf{u}_{ns}|\mathbf x_{ns}^{1:T,i})log\frac{q(\mathbf{u}_s|\mathbf x_s^{t},a^{1:T,i})q(\mathbf{u}_{ns}|\mathbf x_{ns}^{1:T,i})}{p(\mathbf{u}_s|\mathbf x_s^{1:T,i},a^{1:T,i})p(\mathbf{u}_{ns}|\mathbf x_{ns}^{1:T,i})} \nonumber\\
      =&\sum_{i\in I}\sum_{u_s}\sum_{u_{ns}}\beta_i q(\mathbf{u}_s|\mathbf x_s^{1:T,i},a^{1:T,i})q(\mathbf{u}_{ns}|\mathbf x_{ns}^{1:T,i})[log\frac{q(\mathbf{u}_s|\mathbf x_s^{1:T,i},a^{1:T,i})}{p(\mathbf{u}_s|\mathbf x_s^{1:T,i},a^{1:T,i})}+log\frac{q(\mathbf{u}_{ns}|\mathbf x_{ns}^{1:T,i})}{p(\mathbf{u}_{ns}|\mathbf x_{ns}^{1:T,i})}]\nonumber\\
      =&\sum_{i\in I}\sum_{u_s}\sum_{u_{ns}}\beta_i q(\mathbf{u}_s|\mathbf x_s^{1:T,i},a^{1:T,i})q(\mathbf{u}_{ns}|\mathbf x_{ns}^{1:T,i})log\frac{q(\mathbf{u}_s|\mathbf x_s^{1:T,i},a^{1:T,i})}{p(\mathbf{u}_s|\mathbf x_s^{1:T,i},a^{1:T,i})}\nonumber\\&+\sum_{i\in I}\sum_{u_s}\sum_{u_{ns}}\beta_i q(\mathbf{u}_s|\mathbf x_s^{1:T,i},a^{1:T,i})q(\mathbf{u}_{ns}|\mathbf x_{ns}^{1:T,i})log\frac{q(\mathbf{u}_{ns}|\mathbf x_{ns}^{1:T,i})}{p(\mathbf{u}_{ns}|\mathbf x_{ns}^{1:T,i})}\nonumber\\
      =&\sum_{i\in I}\sum_{u_s}\beta_i q(\mathbf{u}_s|\mathbf x_s^{1:T,i},a^{1:T,i})log\frac{q(\mathbf{u}_s|\mathbf x_s^{1:T,i},a^{1:T,i})}{p(\mathbf{u}_s|\mathbf x_s^{1:T,i},a^{1:T,i})}\sum_{u_{ns}}q(\mathbf{u}_{ns}|\mathbf x_{ns}^{1:T,i})\nonumber\\&+\sum_{i\in I}\sum_{u_{ns}}\beta_i q(\mathbf{u}_{ns}|\mathbf x_{ns}^{1:T,i})log\frac{q(\mathbf{u}_{ns}|\mathbf x_{ns}^{1:T,i})}{p(\mathbf{u}_{ns}|\mathbf x_{ns}^{1:T,i})}\sum_{u_{s}}q(\mathbf{u}_s|\mathbf x_s^{1:T,i},a^{1:T,i})\nonumber\\
      =&\sum_{i\in I}\sum_{u_s}\beta_i q(\mathbf{u}_s|\mathbf x_s^{1:T,i},a^{1:T,i})log\frac{q(\mathbf{u}_s|\mathbf x_s^{1:T,i},a^{1:T,i})}{p(\mathbf{u}_s|\mathbf x_s^{1:T,i},a^{1:T,i})}+\sum_{i\in I}\sum_{u_{ns}}\beta_i q(\mathbf{u}_{ns}|\mathbf x_{ns}^{1:T,i})log\frac{q(\mathbf{u}_{ns}|\mathbf x_{ns}^{1:T,i})}{p(\mathbf{u}_{ns}|\mathbf x_{ns}^{1:T,i})}\nonumber\\
      \leq&\sum_{i\in I} \beta_i (\text{KL}(q(\mathbf{u}_s|\mathbf x_s^{1:T,i},a^{1:T,i})||p(\mathbf{u}_s|\mathbf x_{s}^{1:T,i},a^{1:T,i})+\text{KL}(q(\mathbf{u}_{ns}|\mathbf x_{ns}^{1:T,i})||p(\mathbf{u}_{ns}|\mathbf x_{ns}^{1:T,i}))), 
 \end{align}
where the inequality holds for any $I \in \mathcal{I}$, therefore, its infimum can be taken as follows:
\begin{align}
     &\text{KL}(q(\mathbf{u}_s,\mathbf{u}_{ns}|\mathbf x_s^{T+m},a^{T+m},\mathbf x_{ns}^{T+m})||p(\mathbf{u}_s,\mathbf{u}_{ns}|\mathbf x_s^{T+m},a^{T+m},\mathbf x_{ns}^{T+m}))  \nonumber\\ 
     \leq&\inf_{I\in \mathcal{I}} [\sum_{i\in I} \beta_i (\text{KL}(q(\mathbf{u}_s|\mathbf x_s^{1:T,i},a^{1:T,i})||p(\mathbf{u}_s|\mathbf x_{s}^{1:T,i},a^{1:T,i})+\text{KL}(q(\mathbf{u}_{ns}|\mathbf x_{ns}^{1:T,i})||p(\mathbf{u}_{ns}|\mathbf x_{ns}^{1:T,i})))].
\end{align}

\subsection{Derivation of ELBO for DCFDG}
\label{appendix:ELBO}
We assume the prior distribution of latent variables $U_s$ and $U_{ns}$
satisfy Markov property like the following equations:
\begin{align}
    p(\mathbf u_{v1}^t)=p(\mathbf u_{v1}^t|\mathbf u_{v1}^{<t}),
    p(\mathbf u_{v2}^t)=p(\mathbf u_{v2}^t|\mathbf u_{v2}^{<t}).
\end{align}
The joint distribution of data and latent variables is:
\begin{align}
&p(\mathbf x_s^{1:T},\mathbf x_{ns}^{1:T},y^{1:T},\mathbf u_s,\mathbf u_{ns},\mathbf u_{v1}^{1:T},\mathbf u_{v2}^{1:T}|a^{1:T})\nonumber\\=
    &\prod_{t=1}^{T} p(\mathbf x_s^t|\mathbf u_s,\mathbf u_{v1}^t,a^t)
    p(\mathbf x_{ns}^t|\mathbf u_{ns},\mathbf u_{v2}^t)
    p(y^t|\mathbf u_s,\mathbf u_{ns},a^t)\nonumber\\
&\quad\quad p(\mathbf u_s)p(\mathbf u_{ns})p(\mathbf u_{v1}^t)p(\mathbf u_{v2}^t).
\end{align}
According to the causal structure of DCFDG, we can draw the evidence lower bound for $\log p(\mathbf x_s^{1:T},\mathbf x_{ns}^{1:T},y^{1:T}|a^{1:T})$ as:
\begin{align}
      &\log p(\mathbf x_s^{1:T},\mathbf x_{ns}^{1:T},y^{1:T}|a^{1:T}) \nonumber\\
\geq &\mathbb{E}_q log \frac{p(\mathbf x_s^{1:T},\mathbf x_{ns}^{1:T},y^{1:T},\mathbf u_s,\mathbf u_{ns},\mathbf u_{v1}^{1:T},\mathbf u_{v2}^{1:T}|a^{1:T})}{q(\mathbf u_s,\mathbf u_{ns},\mathbf u_{v1}^{1:T},\mathbf u_{v2}^{1:T}|a^{1:T},\mathbf x_s^{1:T},\mathbf x_{ns}^{1:T},y^{1:T})}  \nonumber\\=& \mathbb{E}_q log \frac{\prod_{t=1}^{T} p(\mathbf x_s^t|\mathbf u_s,\mathbf u_{v1}^t,a^t)
    p(\mathbf x_{ns}^t|\mathbf u_{ns},\mathbf u_{v2}^t)
p(y^t|\mathbf u_s,\mathbf u_{ns},a^t)p(\mathbf u_s)p(\mathbf u_{ns})p(\mathbf u_{v1}^t)p(\mathbf u_{v2}^t)}{\prod_{t=1}^{T}q(\mathbf u_s|\mathbf x_s^t,a^t)q(\mathbf u_{ns}|\mathbf x_{ns}^t)q(\mathbf u_{v1}^t|\mathbf u_{v1}^{<t},\mathbf x_s^t)q(\mathbf u_{v2}^t|\mathbf u_{v2}^{<t},\mathbf x_s^t)}
\nonumber\\
=& \mathbb{E}_q \big[-\sum_{t=1}^{T} log \frac{q(\mathbf u_s|\mathbf x_s^t,a^t)}{p(\mathbf u_s)}-\sum_{t=1}^{T} log \frac{q(\mathbf u_{ns}|\mathbf x_{ns}^t)}{p(\mathbf u_{ns})}-\sum_{t=1}^{T} log \frac{q(\mathbf u_{v1}^t|\mathbf u_{v1}^{<t},\mathbf x_s^t)}{p(\mathbf u_{v1}^t)}-\sum_{t=1}^{T} log \frac{q(\mathbf u_{v2}^t|\mathbf u_{v2}^{<t},\mathbf x_s^t)}{p(\mathbf u_{v2}^t)}\nonumber\\
&+\sum_{t=1}^{T} logp(\mathbf x_s^t|\mathbf u_s,\mathbf u_{v1}^t,a^t)
    p(\mathbf x_{ns}^t|\mathbf u_{ns},\mathbf u_{v2}^t)
p(y^t|\mathbf u_s,\mathbf u_{ns},a^t)\big]
\nonumber\\
     \geq& \sum_{t=1}^{T}\{ \mathbb{E}_{q(\mathbf u_s|\mathbf x_s^t,a^t)q(\mathbf u_{v1}^t|\mathbf u_{v1}^{<t},\mathbf x_s^t)}\big[\log{p\left(\mathbf x_s^t|\mathbf u_s,\mathbf u_{v1}^t,a^t\right)}\big] \nonumber\\
     &+ \mathbb{E}_{q(\mathbf u_{ns}|\mathbf x_{ns}^t)q(\mathbf u_{v2}^t|\mathbf u_{v2}^{<t},\mathbf x_{ns}^t)}\big[\log{p\left(\mathbf x_{ns}^t|\mathbf u_{ns},\mathbf u_{v2}^t\right)}\big] \nonumber\\
     &+ \mathbb{E}_{q(\mathbf u_s|\mathbf x_s^t,a^t)q(\mathbf u_{ns}|\mathbf x_{ns}^t)}\big[\log{p\left(y^t|\mathbf u_s,\mathbf u_{ns},a^t\right)}\big] \nonumber\\
    &-KL\big(q(\mathbf u_s|\mathbf x_s^t,a^t)||p(\mathbf u_s)\big)\nonumber\\
    &-KL\big(q(\mathbf u_{ns}|\mathbf x_{ns}^t)||p(\mathbf u_{ns})\big) \nonumber\\
    &-KL\big(q(\mathbf u_{v1}^t|\mathbf u_{v1}^{<t},\mathbf x_s^t)||p(\mathbf u_{v1}^t|\mathbf u_{v1}^{<t})\big) \nonumber\\
    &-KL\big(q(\mathbf u_{v2}^t|\mathbf u_{v2}^{<t},\mathbf x_{ns}^t)||p(\mathbf u_{v2}^t|\mathbf u_{v2}^{<t})\big) \} \nonumber\\
	=&: \text{ELBO} .
\label{eq:ELBO_jensen}
\end{align}
The final greater than or equal to sign is derived using the Jensen's inequality, thus concluding the proof.
\clearpage
\section{Inplementation of Experiments}
\subsection{Visualization of Fair-circle dataset.}
\label{app:faircicle}
\begin{figure}[h]
    \centering
    \includegraphics[width=0.7\columnwidth]{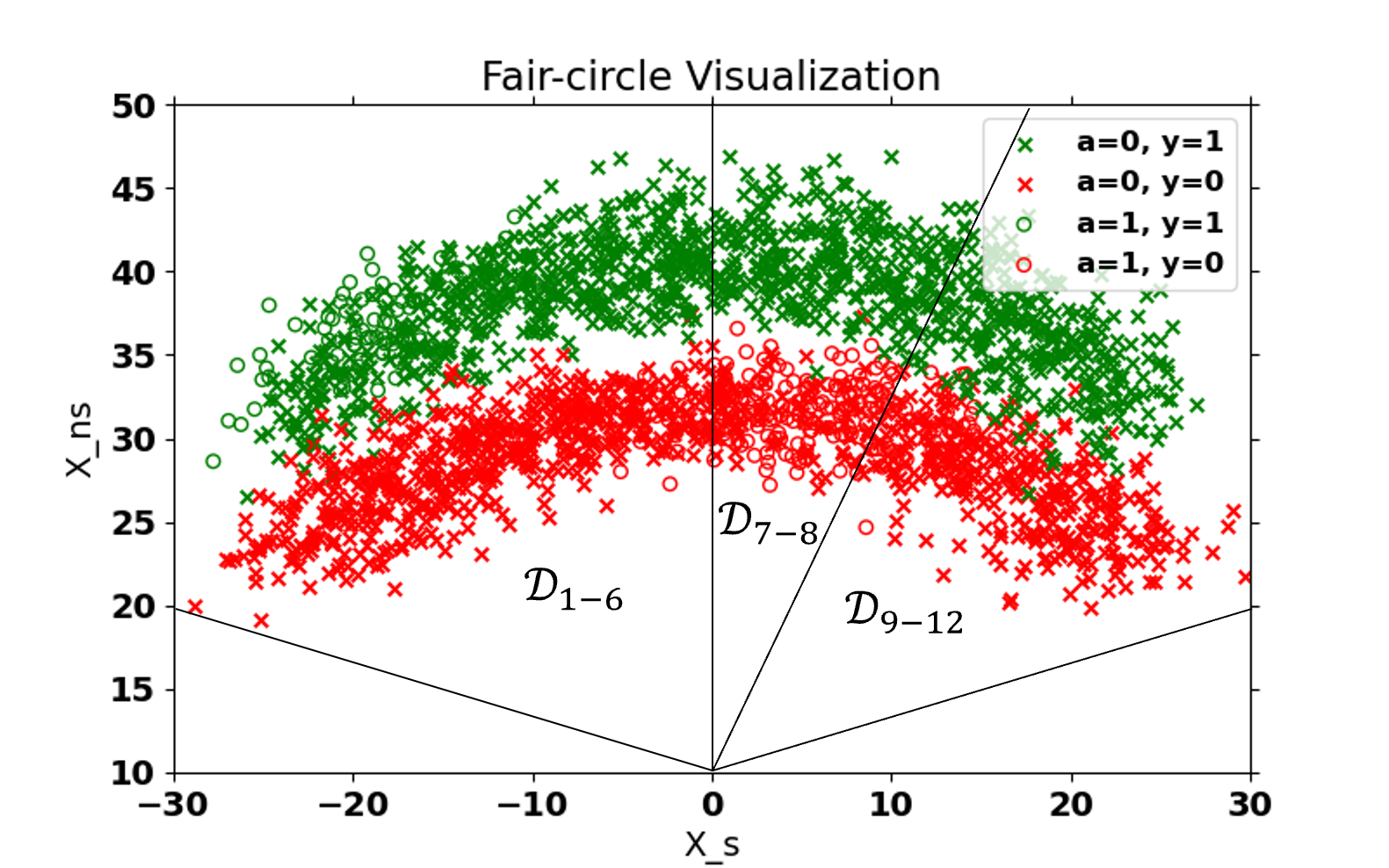}
    \caption{Visualization of Fair-circle dataset. 
From left to right, these are respectively the training set, validation set, and test set.}
    \label{fig:Fair-circle}
\end{figure}
\subsection{Product-Moment Correlation Coefficients (PPMCC) of all three datasets.}
\label{app:PPMCC}
\begin{table}[h]
    \small
    \centering
    \caption{Product-Moment Correlation Coefficients (PPMCC) of all three datasets.}
    \begin{tabular}{l|c|c|c}
        \toprule
        & $X_s$ & $X_{ns}$ & $Y$ \\
        \midrule
        Fair-circle    & $.0910$   & $.0186$ &$.1249$ \\
        Adult   & $.1892$ & $.0597$ &$.2158$\\
        Chicago Crime  & $.0341$   & $.0029$ &$.1355$\\
        \bottomrule
    \end{tabular}
    \label{table:relation}
\end{table}

\subsection{Specific Model Architecture}
\label{app:architecture}
\begin{table*}[h]
\caption{Implementation of Encoder ($E^{s}$ and 
 $E^{ns}$).}
\label{tab:digit_encoder}
\vskip 0.15in
\begin{center}
\begin{tabular}{p{30pt}<{\raggedright} p{125pt}<{\raggedright}}

\toprule
\textbf{\#} & \textbf{Layer} \\
\midrule
1  & Linear(in=$d$, output=128) \\
2  & ReLU \\
3  & Linear(in=128, output=128) \\
4  & ReLU \\
5  & Linear(in=128, output=128) \\
6  & ReLU \\
7  & Linear(in=128, output=$d$) \\
\bottomrule
\end{tabular}
\end{center}

\end{table*}

\begin{table*}[h]
\caption{Implementation of Encoder ($E^{v1}$ and 
 $E^{v2}$).}
\label{tab:digit_encoder}
\vskip 0.15in
\begin{center}
\begin{tabular}{p{25pt}<{\raggedright} p{130pt}<{\raggedright}}
\toprule
\textbf{\#} & \textbf{Layer} \\
\midrule
1  & Linear(in=$d$, output=128) \\
2  & ReLU \\
3  & Linear(in=128, output=128) \\
4  & ReLU \\
5  & Linear(in=128, output=128) \\
6  & ReLU \\
7  & Linear(in=128, output=$d$) \\
\bottomrule
\end{tabular}
\end{center}

\end{table*}

\begin{table*}[h]
\caption{Implementation of Decoder ($D^{s}$ and 
 $D^{ns}$).}
\label{tab:digit_decoder}
\vskip 0.15in
\begin{center}
\begin{tabular}{p{30pt}<{\raggedright} p{125pt}<{\raggedright}}
\toprule
\textbf{\#} & \textbf{Layer} \\
\midrule
1  & Linear(in=$d$, output=16) \\
2  & BatchNorm \\
3  & LeakyReLU(0.2) \\
4  & Linear(in=16, output=64) \\
5  & BatchNorm \\
6  & LeakyReLU(0.2) \\
7  & Linear(in=64, output=128) \\
8 & BatchNorm \\
9  & ReLU \\
10  & Linear(in=128, output=$d$) \\
\bottomrule
\end{tabular}
\end{center}

\end{table*}

\begin{table*}[h]
\caption{Implementation of Classifier $C$.}
\label{tab:digit_decoder}
\vskip 0.15in
\begin{center}
\begin{tabular}{p{25pt}<{\raggedright} p{130pt}<{\raggedright}}
\toprule
\textbf{\#} & \textbf{Layer} \\
\midrule
1  & Linear(in=$d$, output=$d\times4$) \\
2  & ReLU \\
3  & Linear(in=$d\times4$, output=$d$) \\
4  & ReLU \\
5  & Linear(in=$d$, output=$d/4$) \\
6  & ReLU \\
7  & Linear(in=$d/ 4$, output=2) \\
\bottomrule
\end{tabular}
\end{center}

\end{table*}

\begin{table*}[h]
\caption{Implementation of Discriminator $D$.}
\label{tab:digit_decoder}
\vskip 0.15in
\begin{center}
\begin{tabular}{p{25pt}<{\raggedright} p{130pt}<{\raggedright}}
\toprule
\textbf{\#} & \textbf{Layer} \\
\midrule
1  & Linear(in=$d$, output=$128$) \\
2  & ReLU \\
3  & Linear(in=$128$, output=$256$) \\
4  & ReLU \\
5  & Linear(in=$256$, output=$128$) \\
6  & ReLU \\
7  & Linear(in=$128$, output=2) \\
\bottomrule
\end{tabular}
\end{center}

\end{table*}

\clearpage
\newpage
\subsection{CE Values for Ablation Study Outcomes}
\label{app:ablation CE}
\begin{table*}[h]
    \small
    \centering
    \renewcommand{\tabcolsep}{1.5mm}
    \caption{Ablation study results across the two datasets. The results in the table represent the mean values of all test domain outcomes.}
    \begin{tabular}{l| c c c c | c c c c}
        \toprule
         & \multicolumn{4}{c|}{\textbf{Adult}} & \multicolumn{4}{c}{\textbf{Chicago Crime}} \\
        \cmidrule(lr){1-1}\cmidrule(lr){2-5} \cmidrule(lr){6-9} 
        \multirow{2}{*}{Methods}
        & \multicolumn{4}{c|}{CE $\downarrow$ ($\times 10$)}
        & \multicolumn{4}{c}{CE $\downarrow$ ($\times 10$)} \\
        \cmidrule(lr){2-5} \cmidrule(lr){6-9}
         &  $o_{00}$ & $o_{01}$ & $o_{10}$ & $o_{11}$
        &   $o_{00}$ & $o_{01}$ & $o_{10}$ & $o_{11}$ \\
        \midrule
        w/o disentanglement & 1.35   & 0.01
        & 0.53 & 0.50 &1.66 &1.53 &1.69 &1.56\\
        
        w/o fairness loss &3.64  & 1.45
        & 2.57 & 2.91  &0.33 &0.29 &0.31 &0.26\\
        
        \sysname{} (Ours) & \textbf{0.10} & \textbf{0.01}
        & \textbf{0.17} & \textbf{0.26} & \textbf{0.01} & \textbf{0.01} & \textbf{0.01} & \textbf{0.01}\\
        
        \bottomrule
    \end{tabular}
\end{table*}

\subsection{CE Values for Trade-off Outcomes}
\label{app:trade-off CE}
\begin{figure}[h]
  \centering
 \begin{subfigure}[b]{0.24\columnwidth}
    \centering
    \includegraphics[width=\textwidth]{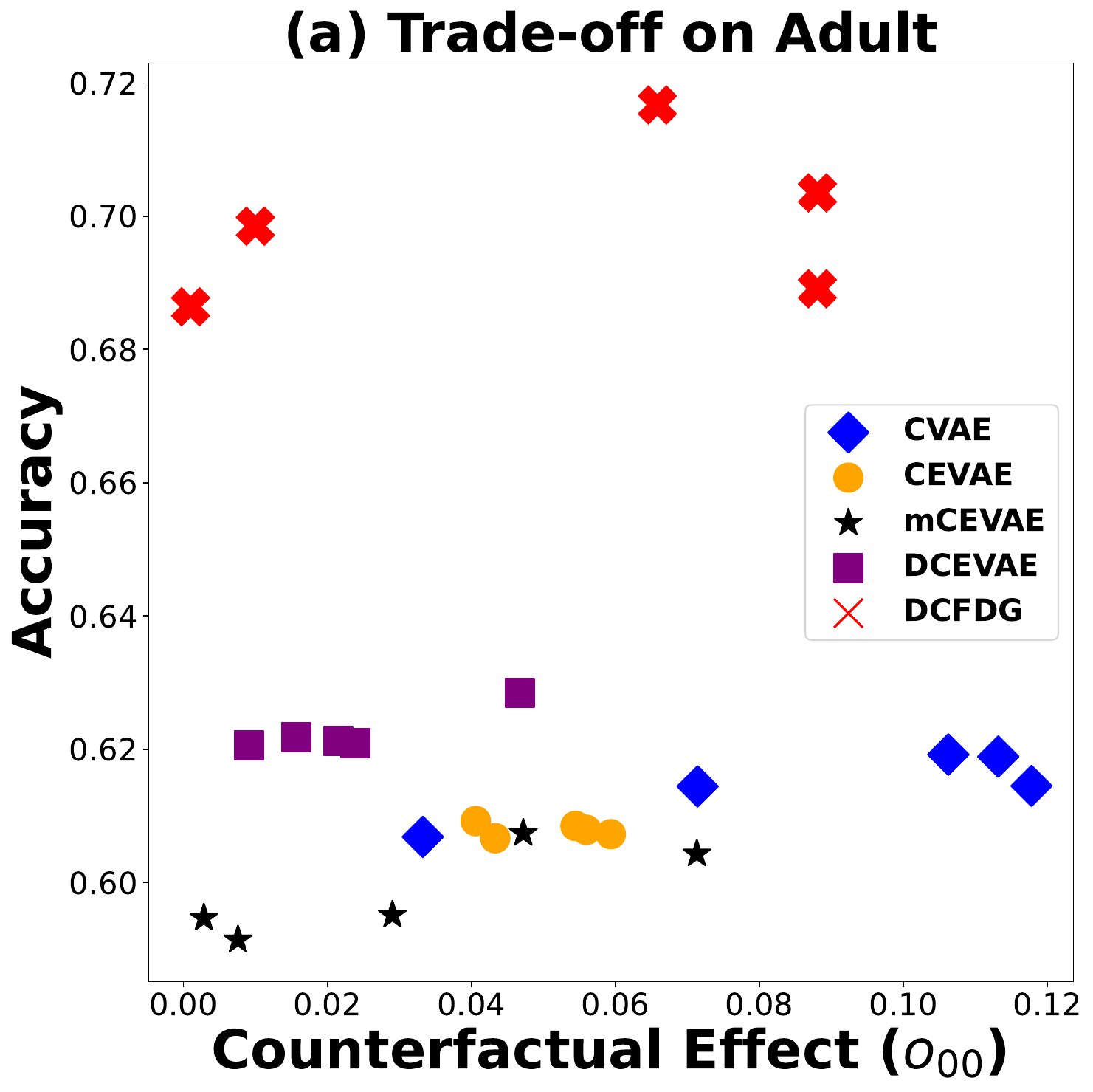}
  \end{subfigure}%
  \begin{subfigure}[b]{0.24\columnwidth}
    \centering
    \includegraphics[width=\textwidth]{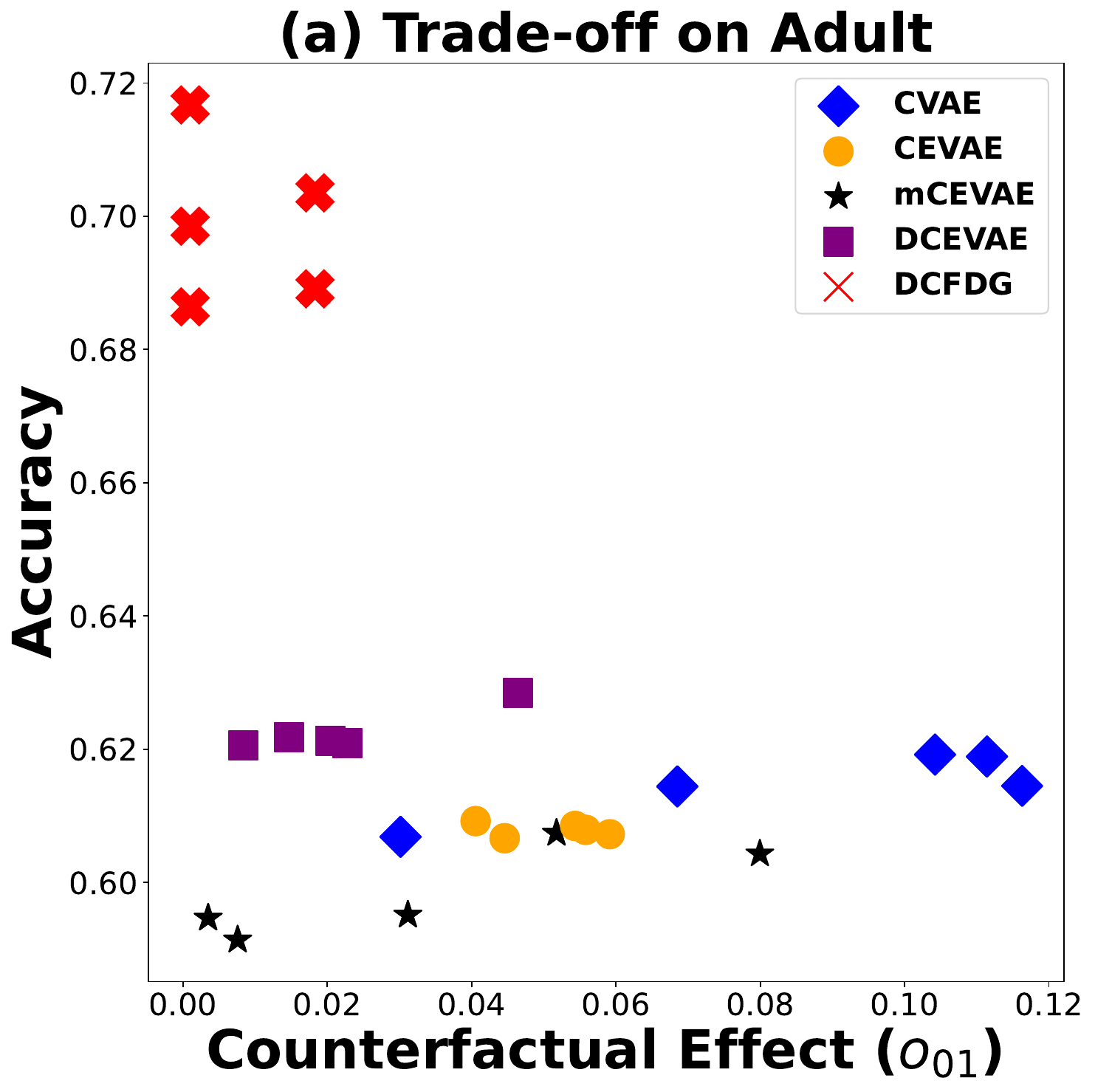}
  \end{subfigure}
  \begin{subfigure}[b]{0.24\columnwidth}
    \centering
    \includegraphics[width=\textwidth]{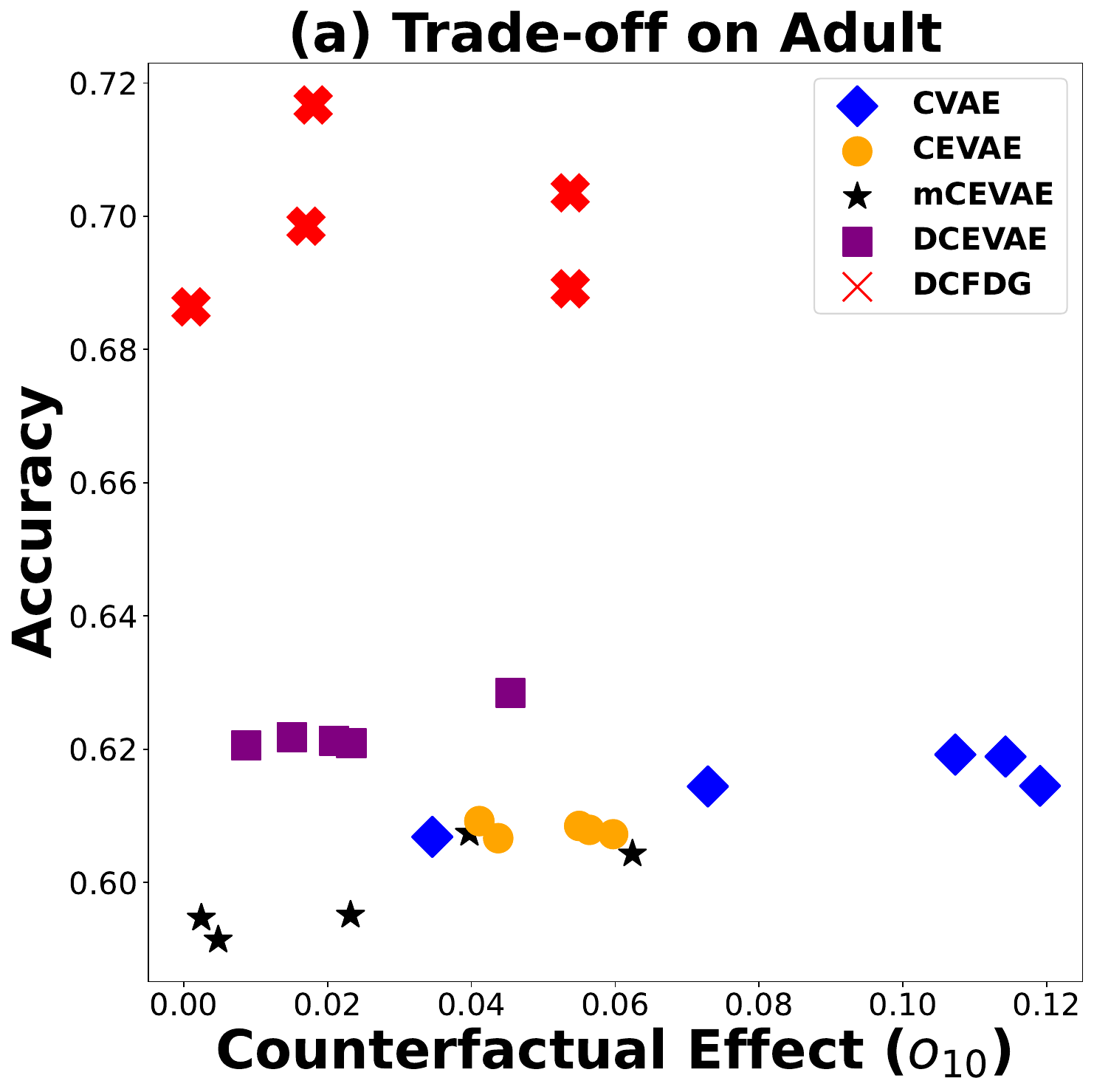}
  \end{subfigure}
  \begin{subfigure}[b]{0.24\columnwidth}
    \centering
    \includegraphics[width=\textwidth]{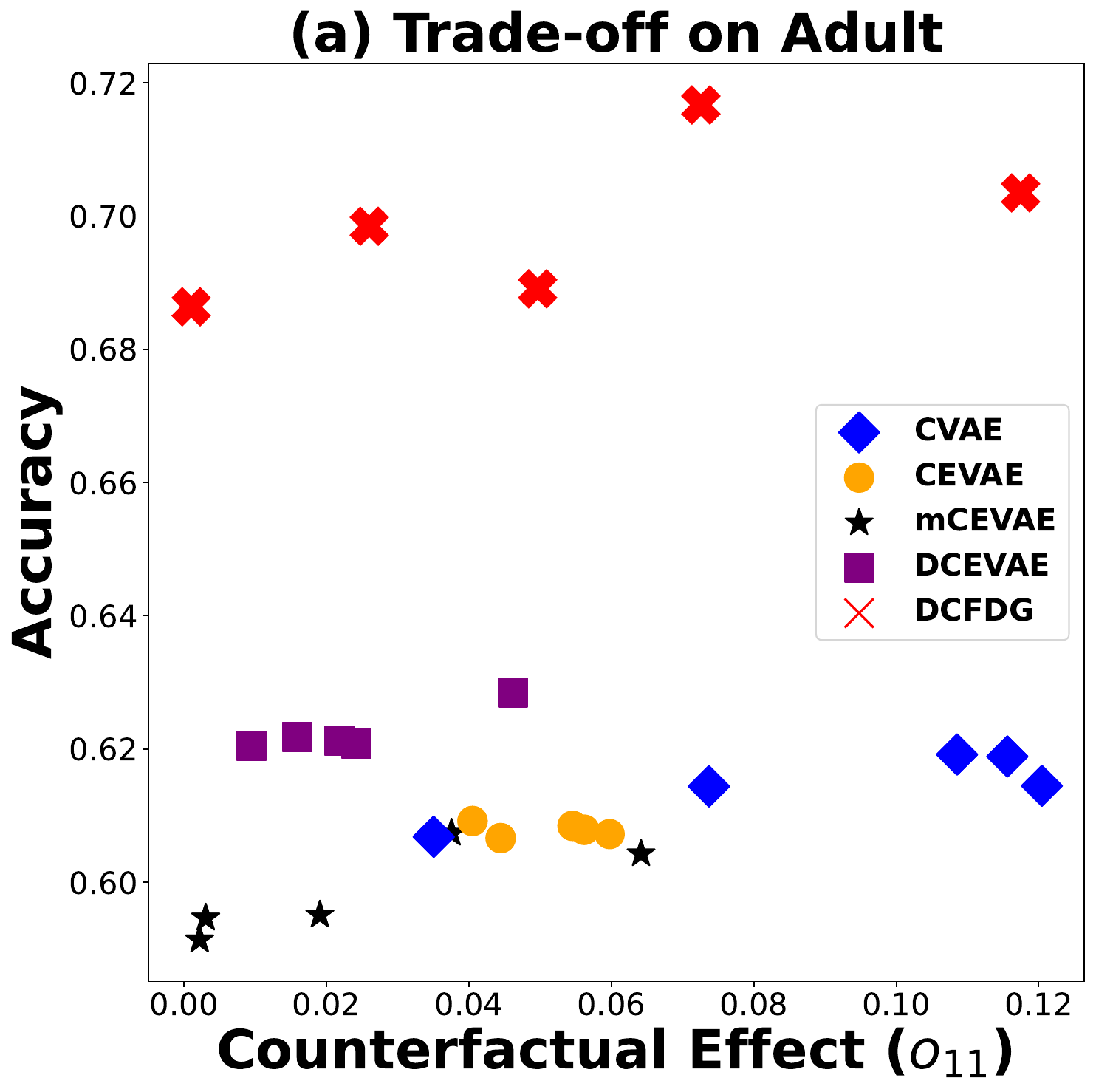}
  \end{subfigure}\\
  \begin{subfigure}[b]{0.24\columnwidth}
    \centering
    \includegraphics[width=\textwidth]{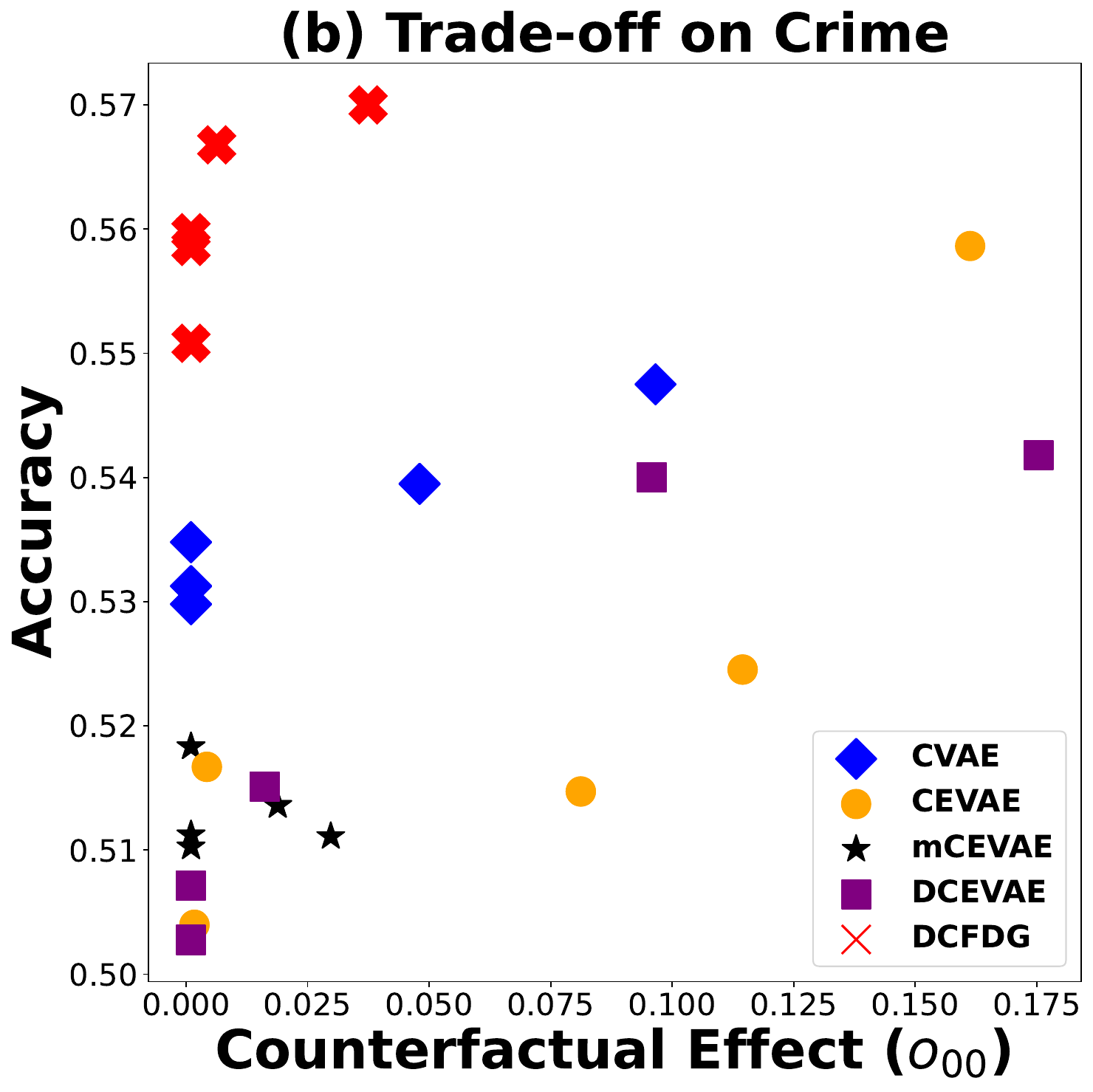}
  \end{subfigure}
  \begin{subfigure}[b]{0.24\columnwidth}
    \centering
    \includegraphics[width=\textwidth]{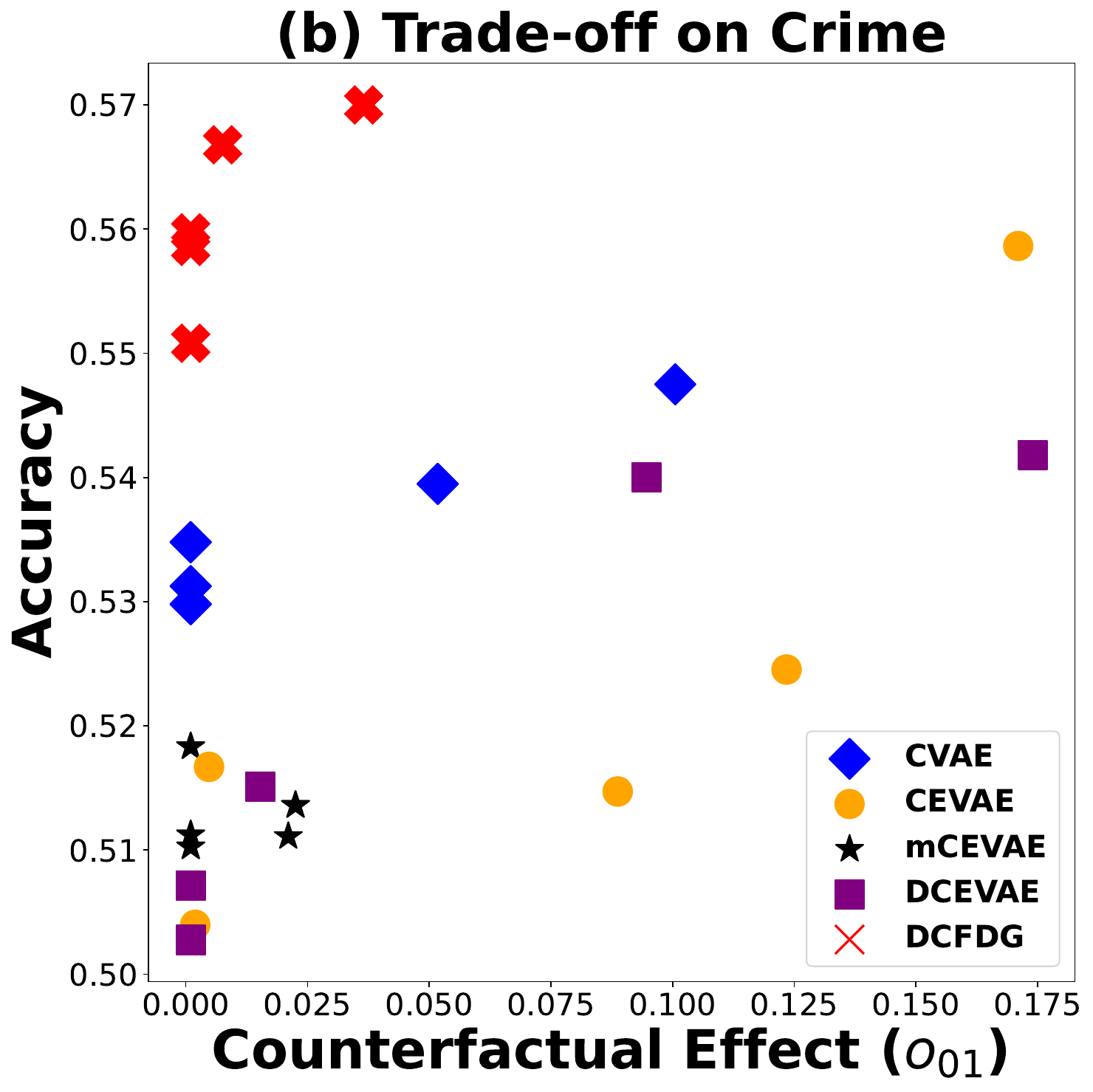}
  \end{subfigure}
  \begin{subfigure}[b]{0.24\columnwidth}
    \centering
    \includegraphics[width=\textwidth]{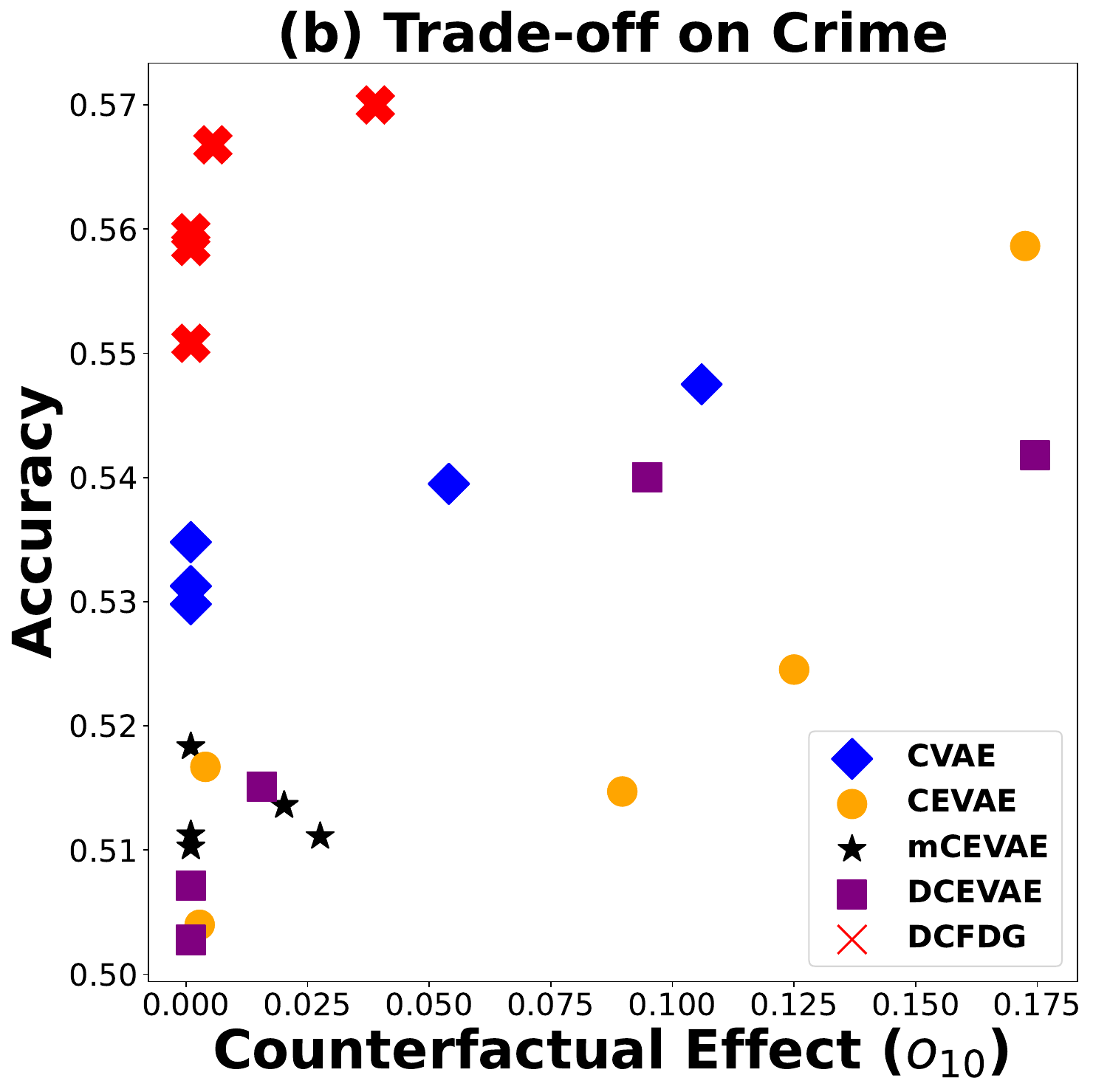}
  \end{subfigure}
  \begin{subfigure}[b]{0.24\columnwidth}
    \centering
    \includegraphics[width=\textwidth]{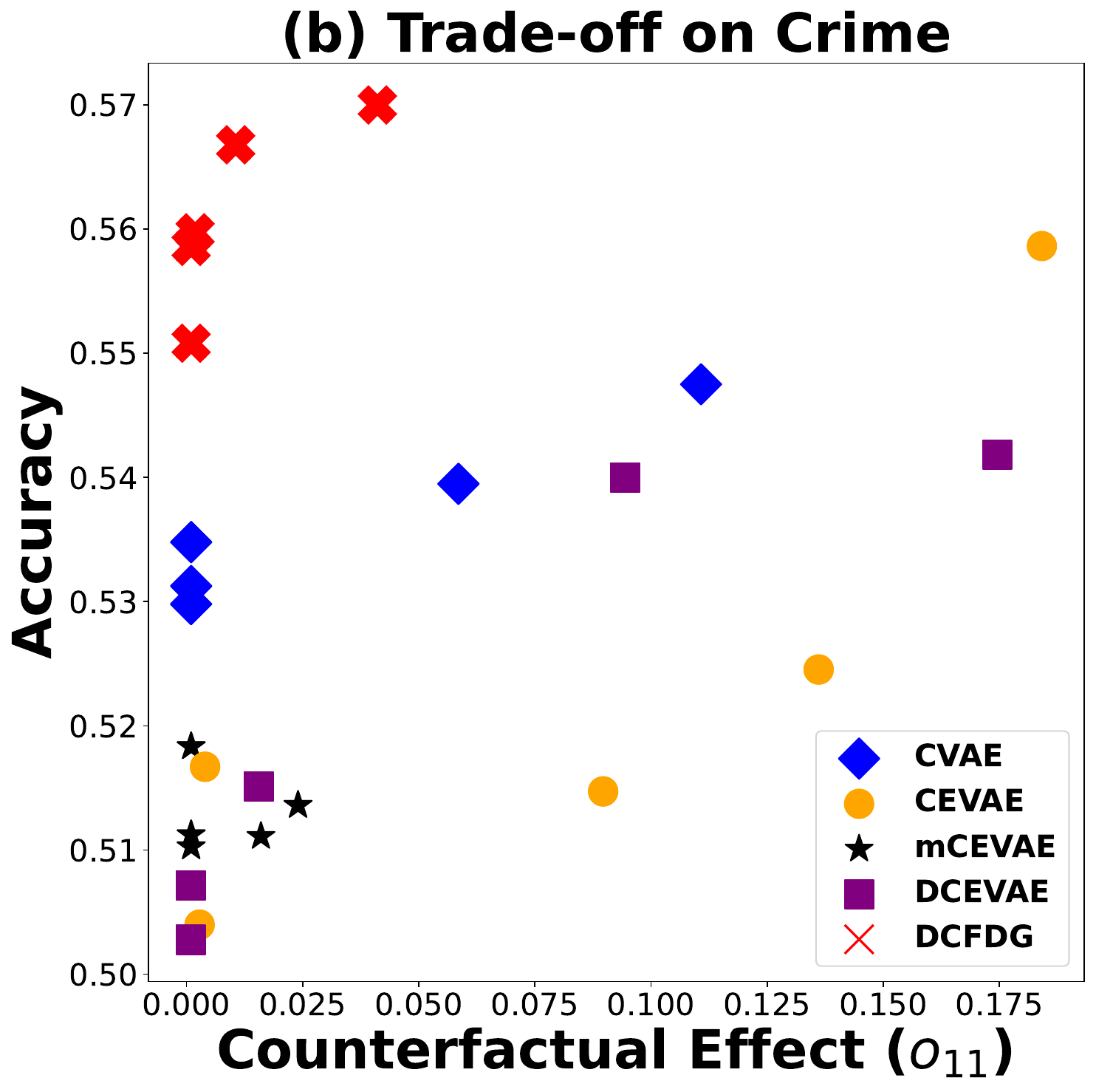}
  \end{subfigure}
  \caption{Fairness-accuracy Trade-off on Adult and Crime. Each baseline is represented by five data points, corresponding to the outcomes under five distinct fairness parameter $\lambda_f$.}
 
\end{figure}
\clearpage
\subsection{Specific Experimental Outcomes Across Each Domain}
\label{app:outcomes across domain}
\subsubsection{Results on the Fair-circle dataset}
\begin{table}[!htbp]
        \caption{Accuracy of the Fair-circle dataset. }
        \centering
        \begin{tabular}{l|cccc}
            \toprule
             & \multicolumn{4}{c}{Accuracy} \\
       
            &T+1 &T+2 &T+3 &T+4\\
            \midrule
            DIVA &96.47$\pm$ 0.17 &77.20$\pm$ 2.36 &52.92$\pm$ 2.01 &50.00$\pm$ 0.00 \\
            LASSE &81.82$\pm$ 5.33 &95.72$\pm$ 3.64 &96.10$\pm$ 3.32 &86.80$\pm$ 2.75 \\
            MMD-LASE &96.10$\pm$3.32 &96.81$\pm$1.55 &82.62$\pm$2.29 &55.65$\pm$1.13 \\
            \midrule
            CVAE &49.88$\pm$0.31 &50.06$\pm$0.23 &50.03$\pm$0.11 &49.98$\pm$0.05 \\
            CEVAE &50.08$\pm$0.27 &49.93$\pm$0.23 &49.96$\pm$0.11 &49.99$\pm$0.09 \\
            mCEVAE &50.24$\pm$0.00 &50.75$\pm$0.18 &64.06$\pm$3.04 &87.06$\pm$1.48 \\
            DCEVAE  &61.99$\pm$1.75 &50.92$\pm$0.51 &50.11$\pm$0.02 &49.95$\pm$0.00 \\
            \sysname{} (Ours) &98.33$\pm$0.30 &98.35$\pm$0.17 &90.88$\pm$0.46 &67.21$\pm$1.46 \\
            \bottomrule
        \end{tabular}
        
        \label{subtable:causal-a}

\end{table}

\begin{table}[!htbp]
        \caption{Total causal effect of the Fair-circle dataset.}
        \centering
        \begin{tabular}{l|cccc}
            \toprule
             & \multicolumn{4}{c}{Total causal effect ($\times 10$)} \\
       
            &T+1 &T+2 &T+3 &T+4\\
            \midrule
            DIVA &1.70$\pm$0.52 &2.00$\pm$0.22 &0.28$\pm$0.19 &0.63$\pm$0.89 \\
            LASSE &4.34$\pm$0.77 &4.72$\pm$0.56 &5.20$\pm$0.88 &5.85$\pm$0.93 \\
            MMD-LASE &0.68$\pm$0.57 &0.23$\pm$0.00 &0.89$\pm$0.20 &1.00$\pm$0.07 \\
            \midrule
            CVAE &0.10$\pm$0.09 &0.15$\pm$0.13 &0.12$\pm$0.10 &0.20$\pm$0.18 \\
            CEVAE &0.15$\pm$0.15 &0.26$\pm$0.16 &0.40$\pm$0.14 &0.55$\pm$0.13 \\
            mCEVAE &0.32$\pm$0.06 &0.27$\pm$0.13 &0.26$\pm$0.15 &0.25$\pm$0.17 \\
            DCEVAE  &0.34$\pm$0.12 &0.22$\pm$0.15 &0.12$\pm$0.14 &0.04$\pm$0.07 \\
            \sysname{} (Ours) &0.07$\pm$0.06 &0.06$\pm$0.03 &0.15$\pm$0.02 &0.20$\pm$0.07 \\
            \bottomrule
        \end{tabular}
        
        \label{subtable:causal-a}

\end{table}

\subsubsection{Results on the Adult dataset}

\begin{table}[!ht]
\caption{Accuracy of the Adult dataset. }
    \begin{center}
        \begin{small}
        \begin{tabular}{l|cccccc}
            \toprule
            & \multicolumn{6}{c}{Accuracy} \\
       
            &T+1 &T+2 &T+3 &T+4 &T+5 &T+6\\
            \midrule
            DIVA &69.82$\pm$1.82 &67.56$\pm$1.71 &67.62$\pm$2.29 &67.78$\pm$2.55 &67.55$\pm$2.11 &67.51$\pm$1.78 \\
            LASSE &60.01$\pm$1.81 &57.34$\pm$1.78 &56.56$\pm$2.5 &56.34$\pm$2.00 &57.16$\pm$1.84 &59.32$\pm$1.85 \\
            MMD-LASE &60.99$\pm$3.56 &60.41$\pm$2.12 &59.73$\pm$1.09 &60.19$\pm$0.43 &59.50$\pm$2.34 &61.21$\pm$2.31 \\
            \midrule
            CVAE &60.60$\pm$0.23 &59.41$\pm$1.65 &58.82$\pm$1.15 &59.52$\pm$1.52 &63.35$\pm$1.28 &69.24$\pm$1.80 \\
            CEVAE &61.02$\pm$0.23 &60.08$\pm$0.28 &59.05$\pm$0.40 &59.79$\pm$0.40 &64.08$\pm$0.42 &70.90$\pm$0.32 \\
            mCEVAE &59.73$\pm$0.71 &59.10$\pm$0.78 &58.1$\pm$0.43 &58.37$\pm$1.47 &62.42$\pm$0.82 &68.53$\pm$1.86 \\
            DCEVAE  &61.27$\pm$0.10 &60.37$\pm$0.07 &59.46$\pm$0.00 &60.11$\pm$0.15 &64.38$\pm$0.02 &71.22$\pm$0.07 \\
            \sysname{} (Ours) &72.71$\pm$0.04 &72.29$\pm$0.04 &68.33$\pm$0.04 &69.64$\pm$0.89 &66.72$\pm$0.04 &69.39$\pm$1.92 \\
            \bottomrule
        \end{tabular}

    \end{small}
    \end{center}

\end{table}
\clearpage
\begin{table}[!ht]
\caption{Total causal effect of the Adult dataset. }
    \begin{center}
        \begin{small}
        \begin{tabular}{l|cccccc}
            \toprule
            & \multicolumn{6}{c}{Total causal effect ($\times 10$)} \\
            &T+1 &T+2 &T+3 &T+4 &T+5 &T+6\\
            \midrule
            DIVA &0.79$\pm$0.11 &0.81$\pm$0.12 &0.86$\pm$0.14 &0.80$\pm$0.11 &0.78$\pm$0.09 &0.80$\pm$0.12 \\
            LASSE &1.92$\pm$0.12 &2.02$\pm$0.06 &1.94$\pm$0.06 &1.96$\pm$0.07 &1.89$\pm$0.01 &1.75$\pm$0.19 \\
            MMD-LASE &1.68$\pm$1.05 &1.64$\pm$0.89 &1.61$\pm$1.01 &1.58$\pm$0.98 &1.55$\pm$1.16 &1.51$\pm$1.20 \\
            \midrule
            CVAE &0.56$\pm$0.47 &0.56$\pm$0.48 &0.57$\pm$0.47 &0.55$\pm$0.48 &0.57$\pm$0.49 &0.56$\pm$0.47 \\
            CEVAE &0.69$\pm$0.27 &0.69$\pm$0.27 &0.69$\pm$0.27 &0.69$\pm$0.27 &0.69$\pm$0.28 &0.69$\pm$0.28 \\
            mCEVAE &0.46$\pm$0.25 &0.45$\pm$0.25 &0.47$\pm$0.28 &0.47$\pm$0.29 &0.46$\pm$0.28 &0.46$\pm$0.28 \\
            DCEVAE  &0.38$\pm$0.05 &0.38$\pm$0.05 &0.38$\pm$0.05 &0.38$\pm$0.05 &0.38$\pm$0.05 &0.38$\pm$0.05 \\
            \sysname{} (Ours) &0.02$\pm$0.02 &0.01$\pm$0.01 &0.01$\pm$0.01 &0.3$\pm$0.05 &0.47$\pm$0.15 &0.52$\pm$0.05 \\
            \bottomrule
        \end{tabular}

    \end{small}
    \end{center}
    
    \end{table}

    \begin{table}[ht]
    \caption{Counterfactual effect of the Adult dataset, where condition $O:=o_{00}$.}
    \begin{center}
        \begin{small}
        \begin{tabular}{l|cccccc}
            \toprule
            & \multicolumn{6}{c}{Counterfactual Effcet: $o_{00}$ ($\times 10$)} \\
       
            &T+1 &T+2 &T+3 &T+4 &T+5 &T+6\\
            \midrule
            DIVA &1.39$\pm$0.74 &1.21$\pm$0.21 &0.50$\pm$0.28 &0.55$\pm$0.43 &0.84$\pm$0.30 &0.78$\pm$0.27 \\
            LASSE &1.93$\pm$0.45 &3.04$\pm$1.89 &3.61$\pm$2.55 &3.63$\pm$2.35 &3.06$\pm$1.69 &2.49$\pm$1.54 \\
            MMD-LASE &0.80$\pm$1.14 &1.52$\pm$2.15 &1.38$\pm$1.96 &1.43$\pm$2.03 &1.33$\pm$1.88 &0.54$\pm$0.77 \\
            \midrule
            CVAE &0.55$\pm$0.47 &0.45$\pm$0.46 &0.50$\pm$0.51 &0.55$\pm$0.50 &0.56$\pm$0.48 &0.55$\pm$0.51 \\
            CEVAE &0.68$\pm$0.27 &0.68$\pm$0.28 &0.68$\pm$0.27 &0.67$\pm$0.25 &0.67$\pm$0.26 &0.67$\pm$0.25 \\
            mCEVAE &0.43$\pm$0.18 &0.44$\pm$0.10 &0.41$\pm$0.13 &0.49$\pm$0.23 &0.45$\pm$0.22 &0.51$\pm$0.20 \\
            DCEVAE  &0.40$\pm$0.06 &0.38$\pm$0.06 &0.38$\pm$0.06 &0.38$\pm$0.06 &0.40$\pm$0.01 &0.38$\pm$0.07 \\
            \sysname{} (Ours) &0.20$\pm$0.28 &0.00$\pm$0.00 &0.30$\pm$0.42 &0.10$\pm$0.15 &0.00$\pm$0.00 &0.00$\pm$0.00 \\
            \bottomrule
        \end{tabular}

    \end{small}
    \end{center}

\end{table}

\begin{table}[htbp]
\caption{Counterfactual effect of the Adult dataset, where condition $O:=o_{01}$.}
    \begin{center}
        \begin{small}
        \begin{tabular}{l|cccccc}
            \toprule
            & \multicolumn{6}{c}{Counterfactual Effcet: $o_{01}$ ($\times 10$)} \\
       
            &T+1 &T+2 &T+3 &T+4 &T+5 &T+6\\
            \midrule
            DIVA &0.61$\pm$0.34 &0.65$\pm$0.39 &0.75$\pm$0.42 &0.41$\pm$0.16 &0.72$\pm$0.48 &0.57$\pm$0.35 \\
            LASSE &3.93$\pm$1.03 &3.07$\pm$0.98 &3.20$\pm$1.39 &4.32$\pm$1.33 &3.87$\pm$1.96 &3.44$\pm$1.47 \\
            MMD-LASE &1.56$\pm$1.59 &1.22$\pm$1.26 &1.39$\pm$1.39 &1.47$\pm$1.33 &1.15$\pm$1.45 &1.29$\pm$1.37 \\
            \midrule
            CVAE &0.53$\pm$0.47 &0.57$\pm$0.45 &0.55$\pm$0.44 &0.53$\pm$0.45 &0.56$\pm$0.46 &0.55$\pm$0.45 \\
            CEVAE &0.69$\pm$0.26 &0.69$\pm$0.26 &0.69$\pm$0.26 &0.69$\pm$0.26 &0.69$\pm$0.26 &0.70$\pm$0.27 \\
            mCEVAE &0.39$\pm$0.05 &0.37$\pm$0.073 &0.34$\pm$0.06 &0.35$\pm$0.05 &0.35$\pm$0.06 &0.34$\pm$0.09 \\
            DCEVAE  &0.38$\pm$0.06 &0.37$\pm$0.06 &0.37$\pm$0.06 &0.38$\pm$0.06 &0.38$\pm$0.05 &0.38$\pm$0.06 \\
            \sysname{} (Ours) &0.05$\pm$0.07 &0.02$\pm$0.04 &0.03$\pm$0.04 &0.03$\pm$0.04 &0.00$\pm$0.00 &0.00$\pm$0.00 \\
            \bottomrule
        \end{tabular}

    \end{small}
    \end{center}

\end{table}

\begin{table}[htbp]
\caption{Counterfactual effect of the Adult dataset, where condition $O:=o_{10}$.}
    \begin{center}
        \begin{small}
        \begin{tabular}{l|cccccc}
            \toprule
            & \multicolumn{6}{c}{Counterfactual Effcet: $o_{10}$ ($\times 10$)} \\
       
            &T+1 &T+2 &T+3 &T+4 &T+5 &T+6\\
            \midrule
            DIVA &0.17$\pm$0.00 &0.19$\pm$0.00 &0.23$\pm$0.00 &0.64$\pm$0.00 &0.49$\pm$0.00 &0.31$\pm$0.00 \\
            LASSE &1.75$\pm$1.11 &1.84$\pm$0.96 &1.65$\pm$0.92 &1.53$\pm$0.68 &1.72$\pm$0.69 &1.74$\pm$1.12 \\
            MMD-LASE &0.96$\pm$1.11 &0.87$\pm$1.12 &1.07$\pm$1.17 &1.08$\pm$1.19 &1.27$\pm$1.23 &1.03$\pm$1.20 \\
            \midrule
            CVAE &0.48$\pm$0.49 &0.55$\pm$0.54 &0.53$\pm$0.50 &0.53$\pm$0.53 &0.51$\pm$0.55 &0.47$\pm$0.45 \\
            CEVAE &0.70$\pm$0.29 &0.69$\pm$0.29 &0.69$\pm$0.28 &0.69$\pm$0.29 &0.69$\pm$0.29 &0.70$\pm$0.30 \\
            mCEVAE &0.51$\pm$0.34 &0.48$\pm$0.31 &0.52$\pm$0.39 &0.50$\pm$0.38 &0.52$\pm$0.37 &0.45$\pm$0.27 \\
            DCEVAE  &0.37$\pm$0.05 &0.39$\pm$0.06 &0.38$\pm$0.05 &0.39$\pm$0.05 &0.38$\pm$0.05 &0.38$\pm$0.06 \\
            \sysname{} (Ours) &0.18$\pm$0.26 &0.24$\pm$0.34 &0.05$\pm$0.07 &0.13$\pm$0.17 &0.21$\pm$0.29 &0.22$\pm$0.30 \\
            \bottomrule
        \end{tabular}

    \end{small}
    \end{center}

\end{table}
\begin{table}[htbp]
\caption{Counterfactual effect of the Adult dataset, where condition $O:=o_{11}$.}
    \begin{center}
        \begin{small}
        \begin{tabular}{l|cccccc}
            \toprule
            & \multicolumn{6}{c}{Counterfactual Effcet: $o_{11}$ ($\times 10$)} \\
       
            &T+1 &T+2 &T+3 &T+4 &T+5 &T+6\\
            \midrule
            DIVA &0.84$\pm$0.14 &0.86$\pm$0.12 &0.93$\pm$0.13 &0.86$\pm$0.13 &0.80$\pm$0.07 &0.86$\pm$0.11 \\
            LASSE &1.65$\pm$0.37 &1.83$\pm$0.37 &1.72$\pm$0.27 &1.67$\pm$0.36 &1.66$\pm$0.26 &1.53$\pm$0.06 \\
            MMD-LASE &1.78$\pm$0.94 &1.75$\pm$0.76 &1.69$\pm$0.87 &1.64$\pm$0.86 &1.63$\pm$1.04 &1.61$\pm$1.18 \\
            \midrule
            CVAE &0.57$\pm$0.47 &0.57$\pm$0.47 &0.57$\pm$0.47 &0.56$\pm$0.47 &0.57$\pm$0.48 &0.57$\pm$0.47 \\
            CEVAE &0.69$\pm$0.27 &0.69$\pm$0.27 &0.69$\pm$0.27 &0.69$\pm$0.27 &0.69$\pm$0.28 &0.69$\pm$0.28 \\
            mCEVAE &0.47$\pm$0.29 &0.47$\pm$0.30 &0.48$\pm$0.31 &0.48$\pm$0.34 &0.47$\pm$0.32 &0.48$\pm$0.33 \\
            DCEVAE  &0.38$\pm$0.05 &0.38$\pm$0.05 &0.38$\pm$0.05 &0.38$\pm$0.05 &0.38$\pm$0.05 &0.38$\pm$0.05 \\
            \sysname{} (Ours) &0.00$\pm$0.00 &0.00$\pm$0.00 &0.00$\pm$0.00 &0.36$\pm$0.04 &0.57$\pm$0.16 &0.63$\pm$0.08 \\
            \bottomrule
        \end{tabular}

    \end{small}
    \end{center}

\end{table}

\clearpage

\subsubsection{Results on the Chicago Crime dataset}
\begin{table}[htbp]
\caption{Accuracy of the Chicago Crime dataset. }
    \begin{center}
        \begin{small}
        \begin{tabular}{l|cccccc}
            \toprule
            & \multicolumn{6}{c}{Accuracy} \\
       
            &T+1 &T+2 &T+3 &T+4 &T+5 &T+6\\
            \midrule
            DIVA &59.65$\pm$0.73 &54.41$\pm$0.37 &51.97$\pm$1.94 &53.85$\pm$1.89 &58.15$\pm$1.79 &59.02$\pm$2.35 \\
            LASSE &52.74$\pm$0.73 &51.33$\pm$0.27 &50.56$\pm$0.17 &53.35$\pm$0.06 &56.64$\pm$1.22 &57.68$\pm$1.15 \\
            MMD-LASE &55.18$\pm$0.27 &53.58$\pm$0.94 &48.27$\pm$3.28 &53.31$\pm$4.40 &56.54$\pm$0.80 &56.10$\pm$0.59 \\
            \midrule
            CVAE &53.63$\pm$2.45 &52.34$\pm$0.82 &51.32$\pm$2.82 &53.15$\pm$1.44 &58.91$\pm$0.79 &51.26$\pm$2.60 \\
            CEVAE &53.35$\pm$1.08 &53.17$\pm$5.33 &52.91$\pm$5.05 &54.33$\pm$2.62 &56.59$\pm$4.85 &51.55$\pm$0.84 \\
            mCEVAE &54.66$\pm$2.86 &50.39$\pm$0.14 &48.38$\pm$0.34 &50.19$\pm$1.03 &55.81$\pm$0.41 &51.55$\pm$0.84 \\
            DCEVAE  &53.86$\pm$0.14 &47.24$\pm$0.03 &43.58$\pm$2.03 &47.04$\pm$0.03 &56.37$\pm$1.20  &59.66$\pm$4.30 \\
            \sysname{} (Ours) &58.47$\pm$0.10 &57.01$\pm$0.55 &55.28$\pm$0.20 &54.34$\pm$0.55 &56.10$\pm$0.87 &54.37$\pm$0.38 \\
            \bottomrule
        \end{tabular}

    \end{small}
    \end{center}

\end{table}
\begin{table}[htbp]
\caption{Total causal effect of the Chicago Crime dataset.}
    \begin{center}
        \begin{small}
        \begin{tabular}{l|cccccc}
            \toprule
            & \multicolumn{6}{c}{Total causal effect ($\times 10$)} \\
       
            &T+1 &T+2 &T+3 &T+4 &T+5 &T+6\\
            \midrule
            DIVA &1.42$\pm$0.14 &1.60$\pm$0.10 &1.61$\pm$0.26 &1.71$\pm$0.04 &1.99$\pm$0.21 &1.73$\pm$0.21 \\
            LASSE &0.63$\pm$0.16 &0.64$\pm$0.34 &0.73$\pm$0.23 &0.94$\pm$0.30 &0.91$\pm$0.54 &1.25$\pm$0.89 \\
            MMD-LASE &0.35$\pm$0.23 &0.29$\pm$0.05 &0.33$\pm$0.15 &0.40$\pm$0.20 &0.37$\pm$0.15 &0.39$\pm$0.25 \\
            \midrule
            CVAE &0.71$\pm$0.01 &0.74$\pm$0.02 &0.73$\pm$0.01 &0.70$\pm$0.01 &0.73$\pm$0.02 &0.72$\pm$0.03 \\
            CEVAE &0.41$\pm$0.19 &0.41$\pm$0.19 &0.43$\pm$0.18 &0.44$\pm$0.19 &0.41$\pm$0.18 &0.42$\pm$0.21 \\
            mCEVAE &0.01$\pm$0.00 &0.01$\pm$0.00 &0.01$\pm$0.00 &0.01$\pm$0.00 &0.01$\pm$0.00 &0.01$\pm$0.00 \\
            DCEVAE  &0.44$\pm$0.05 &0.46$\pm$0.05 &0.45$\pm$0.04 &0.44$\pm$0.05 &0.44$\pm$0.06 &0.42$\pm$0.04 \\
            \sysname{} (Ours) &0.01$\pm$0.00 &0.01$\pm$0.00 &0.01$\pm$0.00 &0.01$\pm$0.00 &0.01$\pm$0.00 &0.01$\pm$0.00 \\
            \bottomrule
        \end{tabular}

    \end{small}
    \end{center}

\end{table}

    \begin{table}[htbp]
    \caption{Counterfactual effect of the Chicago Crime dataset, where condition $O:=o_{00}$.}
    \begin{center}
        \begin{small}
        \begin{tabular}{l|cccccc}
            \toprule
            & \multicolumn{6}{c}{Counterfactual Effcet: $o_{00}$ ($\times 10$)} \\
       
            &T+1 &T+2 &T+3 &T+4 &T+5 &T+6\\
            \midrule
            DIVA &1.39$\pm$0.15 &1.69$\pm$0.13 &1.53$\pm$0.23 &1.91$\pm$0.32 &1.97$\pm$0.12 &1.60$\pm$0.28 \\
            LASSE &0.40$\pm$0.16 &0.35$\pm$0.46 &0.75$\pm$0.31 &0.81$\pm$0.59 &0.92$\pm$0.51 &1.37$\pm$0.96 \\
            MMD-LASE &0.23$\pm$0.16 &0.33$\pm$0.12 &0.29$\pm$0.11 &0.36$\pm$0.18 &0.34$\pm$0.03 &0.36$\pm$0.03 \\
            \midrule
            CVAE &0.66$\pm$0.00 &0.67$\pm$0.04 &0.64$\pm$0.00 &0.66$\pm$0.01 &0.70$\pm$0.01 &0.66$\pm$0.05 \\
            CEVAE &0.39$\pm$0.19 &0.38$\pm$0.19 &0.41$\pm$0.18 &0.42$\pm$0.21 &0.40$\pm$0.19 &0.39$\pm$0.21 \\
            mCEVAE  &0.01$\pm$0.00 &0.01$\pm$0.00 &0.01$\pm$0.00 &0.01$\pm$0.00 &0.01$\pm$0.00 &0.01$\pm$0.00\\
            DCEVAE  &0.47$\pm$0.06 &0.51$\pm$0.07 &0.49$\pm$0.07 &0.47$\pm$0.06 &0.48$\pm$0.08 &0.46$\pm$0.05 \\
            \sysname{} (Ours) &0.01$\pm$0.01 &0.02$\pm$0.03 &0.05$\pm$0.00 &0.02$\pm$0.03 &0.01$\pm$0.01 & 0.01$\pm$0.01\\
            \bottomrule
        \end{tabular}

    \end{small}
    \end{center}

\end{table}

\begin{table}[htbp]
\caption{Counterfactual effect of the Chicago Crime dataset, where condition $O:=o_{01}$.}
    \begin{center}
        \begin{small}
        \begin{tabular}{l|cccccc}
            \toprule
            & \multicolumn{6}{c}{Counterfactual Effcet: $o_{01}$ ($\times 10$)} \\
       
            &T+1 &T+2 &T+3 &T+4 &T+5 &T+6\\
            \midrule
            DIVA &1.13$\pm$0.19 &1.58$\pm$0.27 &1.17$\pm$0.21 &1.52$\pm$0.09 &1.76$\pm$0.11 &1.57$\pm$0.15 \\
            LASSE &0.61$\pm$0.24 &0.74$\pm$0.23 &0.97$\pm$0.57 &0.97$\pm$0.21 &0.97$\pm$0.70 &1.32$\pm$0.80 \\
            MMD-LASE &0.58$\pm$0.24 &0.28$\pm$0.01 &0.43$\pm$0.19 &0.39$\pm$0.24 &0.31$\pm$0.30 &0.48$\pm$0.48 \\
            \midrule
            CVAE &0.69$\pm$0.01 &0.71$\pm$0.00 &0.74$\pm$0.00 &0.66$\pm$0.01 &0.70$\pm$0.00 &0.72$\pm$0.01 \\
            CEVAE &0.41$\pm$0.22 &0.42$\pm$0.19 &0.43$\pm$0.19 &0.46$\pm$0.21 &0.41$\pm$0.20 &0.43$\pm$0.22 \\
            mCEVAE  &0.01$\pm$0.00 &0.01$\pm$0.00 &0.01$\pm$0.00 &0.01$\pm$0.00 &0.01$\pm$0.00 &0.01$\pm$0.00\\
            DCEVAE  &0.45$\pm$0.05 &0.46$\pm$0.06 &0.46$\pm$0.04 &0.45$\pm$0.05 &0.45$\pm$0.05 &0.43$\pm$0.03 \\
            \sysname{} (Ours) &0.02$\pm$0.03 &0.01$\pm$0.02 &0.01$\pm$0.02 &0.02$\pm$0.03 &0.00$\pm$0.00 &0.02$\pm$0.03 \\
            \bottomrule
        \end{tabular}

    \end{small}
    \end{center}

\end{table}

\begin{table}[htbp]
\caption{Counterfactual effect of the Chicago Crime dataset, where condition $O:=o_{10}$.}
    \begin{center}
        \begin{small}
        \begin{tabular}{l|cccccc}
            \toprule
            & \multicolumn{6}{c}{Counterfactual Effcet: $o_{10}$ ($\times 10$)} \\
       
            &T+1 &T+2 &T+3 &T+4 &T+5 &T+6\\
            \midrule
            DIVA &1.77$\pm$0.18 &1.63$\pm$0.15 &1.71$\pm$0.25 &1.93$\pm$0.07 &2.21$\pm$0.25 &1.77$\pm$0.17 \\
            LASSE &0.65$\pm$0.05 &0.75$\pm$0.34 &0.75$\pm$0.13 &0.89$\pm$0.31 &1.06$\pm$0.59 &1.28$\pm$0.93 \\
            MMD-LASE &0.24$\pm$0.14 &0.25$\pm$0.01 &0.31$\pm$0.20 &0.31$\pm$0.10 &0.58$\pm$0.35 &0.45$\pm$0.26 \\
            \midrule
            CVAE &0.74$\pm$0.02 &0.75$\pm$0.03 &0.73$\pm$0.03 &0.72$\pm$0.04 &0.75$\pm$0.05 &0.722$\pm$0.05 \\
            CEVAE &0.40$\pm$0.20 &0.41$\pm$0.19 &0.42$\pm$0.18 &0.44$\pm$0.20 &0.42$\pm$0.19 &0.42$\pm$0.21 \\
            mCEVAE  &0.01$\pm$0.00 &0.01$\pm$0.00 &0.01$\pm$0.00 &0.01$\pm$0.00 &0.01$\pm$0.00 &0.01$\pm$0.00\\
            DCEVAE  &0.44$\pm$0.06 &0.45$\pm$0.07 &0.43$\pm$0.08 &0.43$\pm$0.08 &0.45$\pm$0.07 &0.43$\pm$0.06 \\
            \sysname{} (Ours) &0.01$\pm$0.02 &0.01$\pm$0.00 &0.01$\pm$0.15 &0.03$\pm$0.05 &0.00$\pm$0.00 &0.01$\pm$0.00 \\
            \bottomrule
        \end{tabular}

    \end{small}
    \end{center}

\end{table}

\begin{table}[htbp]
\caption{Counterfactual effect of the Chicago Crime dataset, where condition $O:=o_{11}$.}
    \begin{center}
        \begin{small}
        \begin{tabular}{l|cccccc}
            \toprule
            & \multicolumn{6}{c}{Counterfactual Effcet: $o_{11}$ ($\times 10$)} \\
       
            &T+1 &T+2 &T+3 &T+4 &T+5 &T+6\\
            \midrule
            DIVA &1.38$\pm$0.18 &1.50$\pm$0.08 &2.08$\pm$0.40 &1.49$\pm$0.15 &2.03$\pm$0.37 &2.03$\pm$0.38 \\
            LASSE &0.82$\pm$0.18 &0.66$\pm$0.34 &0.38$\pm$0.13 &1.08$\pm$0.16 &0.67$\pm$0.32 &1.03$\pm$0.88 \\
            MMD-LASE &0.33$\pm$0.38 &0.30$\pm$0.13 &0.28$\pm$0.06 &0.54$\pm$0.28 &0.23$\pm$0.11 &0.21$\pm$0.20 \\
            \midrule
            CVAE &0.73$\pm$0.00 &0.80$\pm$0.01 &0.81$\pm$0.01 &0.75$\pm$0.01 &0.76$\pm$0.02 &0.76$\pm$0.00 \\
            CEVAE &0.43$\pm$0.16 &0.43$\pm$0.17 &0.45$\pm$0.17 &0.45$\pm$0.15 &0.41$\pm$0.15 &0.46$\pm$0.18 \\
            mCEVAE &0.01$\pm$0.00 &0.01$\pm$0.00 &0.01$\pm$0.00 &0.01$\pm$0.00 &0.01$\pm$0.00 &0.01$\pm$0.00\\
            DCEVAE  &0.39$\pm$0.02 &0.40$\pm$0.02 &0.39$\pm$0.01 &0.39$\pm$0.02 &0.38$\pm$0.02 &0.36$\pm$0.01 \\
            \sysname{} (Ours) &0.04$\pm$0.05 &0.01$\pm$0.01 &0.01$\pm$0.01 &0.02$\pm$0.03 &0.01$\pm$0.01 &0.01$\pm$0.01 \\
            \bottomrule
        \end{tabular}

    \end{small}
    \end{center}

\end{table}
\clearpage


\end{document}